%% file: camera_ready.tex
\documentclass{article}

\PassOptionsToPackage{numbers, sort&compress}{natbib}

 \usepackage[main, final]{neurips_2026}

\usepackage[utf8]{inputenc}
\usepackage[T1]{fontenc}

\usepackage{url}
\usepackage{booktabs}
\usepackage{amsfonts}
\usepackage{nicefrac}
\usepackage{microtype}
\usepackage{xcolor}
\usepackage{caption}
\usepackage{xspace}
\usepackage{amssymb}
\usepackage{amsmath}
\usepackage{algorithm}
\usepackage{algpseudocode}
\usepackage{multirow}
\usepackage{wrapfig}
\usepackage{makecell}
\usepackage[table]{xcolor}
\usepackage[normalem]{ulem}
\usepackage{graphicx}
\usepackage{uoftcolors}

\usepackage[pagebackref]{hyperref}
\usepackage[nameinlink]{cleveref}

\hypersetup{colorlinks=true,allcolors=uoftsecondaryblue}

\title{Good Token Hunting: A Hitchhiker's Guide to Token Selection for Visual Geometry Transformers}

\author{
\vspace{2mm} 
        Shuhong Zheng$^{1,2}$\hspace{6mm}
        Michael Oechsle$^{3}$ \hspace{6mm}
        Erik Sandström$^{3}$ 
        \\
        \textbf{
        Marie-Julie Rakotosaona$^{3}$ \hspace{5mm}
         Federico Tombari$^{3,4\dagger}$ \hspace{5mm}
Igor Gilitschenski$^{1,2\dagger}$
    }
    \vspace{2mm}
    \\
    $^{1}$University of Toronto
    \hspace{3mm}  
    $^{2}$Vector Institute 
    \hspace{3mm}  
    $^{3}$Google 
    \hspace{3mm}  
    $^{4}$Technical University of Munich
    \vspace{2mm}
    \\
    {\texttt{\{shuhong, gilitschenski\}@cs.toronto.edu}}
    \vspace{2mm}
    \\{\texttt{\{michaeloechsle, sandstrom, mrakotosaona, tombari\}@google.com}
    }
    }

\newcommand{\methodname}[0]{\textit{GoToHunt}\xspace} 

\newcommand{\pithree}{\pi^3}

\newcommand{\eg}{\textit{e.g.}}

\newcommand{\etc}{\textit{etc.}}

\DeclareMathOperator*{\argmin}{arg\,min}

\begin{document}

\maketitle

\renewcommand{\thefootnote}{$\dagger$}
\footnotetext{Joint Advising}
\renewcommand{\thefootnote}{\arabic{footnote}}

\newif\ifshownotes


\input{secs/0_abstract_arxiv}    
\input{secs/1_intro}
\input{secs/2_related}
\input{secs/3_method_arxiv}

\input{secs/4_comparison}

\input{secs/5_discussion}

\input{secs/6_conclusion}

{
    \small
    \bibliographystyle{ieeenat_fullname}
\bibliography{bib/cut3r_related,bib/video_related,bib/dataset_related}
}

\newpage
\input{checklist.tex}


\newpage

\appendix

\renewcommand\thesection{\Alph{section}}
\renewcommand{\thefigure}{\Alph{figure}}
\renewcommand{\thetable}{\Alph{table}}
\renewcommand\thealgorithm{\Alph{algorithm}}
\renewcommand\theequation{\Alph{equation}}

\setcounter{section}{0}
\setcounter{equation}{0}
\setcounter{figure}{0}
\setcounter{table}{0}
\setcounter{algorithm}{0}

\input{secs/X_supp}

\end{document}

%% file: secs/0_abstract_arxiv.tex
\begin{abstract}
Visual geometry transformers have become powerful architectures for multi-view 3D reconstruction, enabling joint prediction of multiple 3D attributes in a feed-forward manner. However, their computational cost grows quadratically with the input sequence length due to the global attention layers inside these models. This limits both their scalability and efficiency. In this work, we address this challenge with a simple yet general strategy: restricting the number of key/value tokens that each query interacts with during global attention. To achieve effective token selection, we introduce a two-stage framework. First, an inter-frame selection step operates at the frame level to identify frames that should be preserved. Second, an intra-frame selection step further discards more redundant tokens within the selected frames. Our analysis highlights the advantage of a diversity-based strategy for inter-frame selection, which ensures broad coverage of the scene. For intra-frame selection, we show that layer-aware sparsification is necessary, with the selection process guided by the entropy of the global attention pattern. Our approach offers a superior speed-accuracy trade-off compared to existing solutions. Extensive experiments show that it accelerates visual geometry transformers by over 85\% for scenes with 500 images while maintaining, or even improving, baseline performance, which hints that how our token selection strategy can play a crucial role in future applications of visual geometry transformers. Our project website is available at \url{https://zsh2000.github.io/good-token-hunting.github.io/}.
\end{abstract}

%% file: secs/1_intro.tex
\section{Introduction}
\label{sec:intro}

\begin{figure*}[t]
    \centering
    \includegraphics[width=\linewidth]{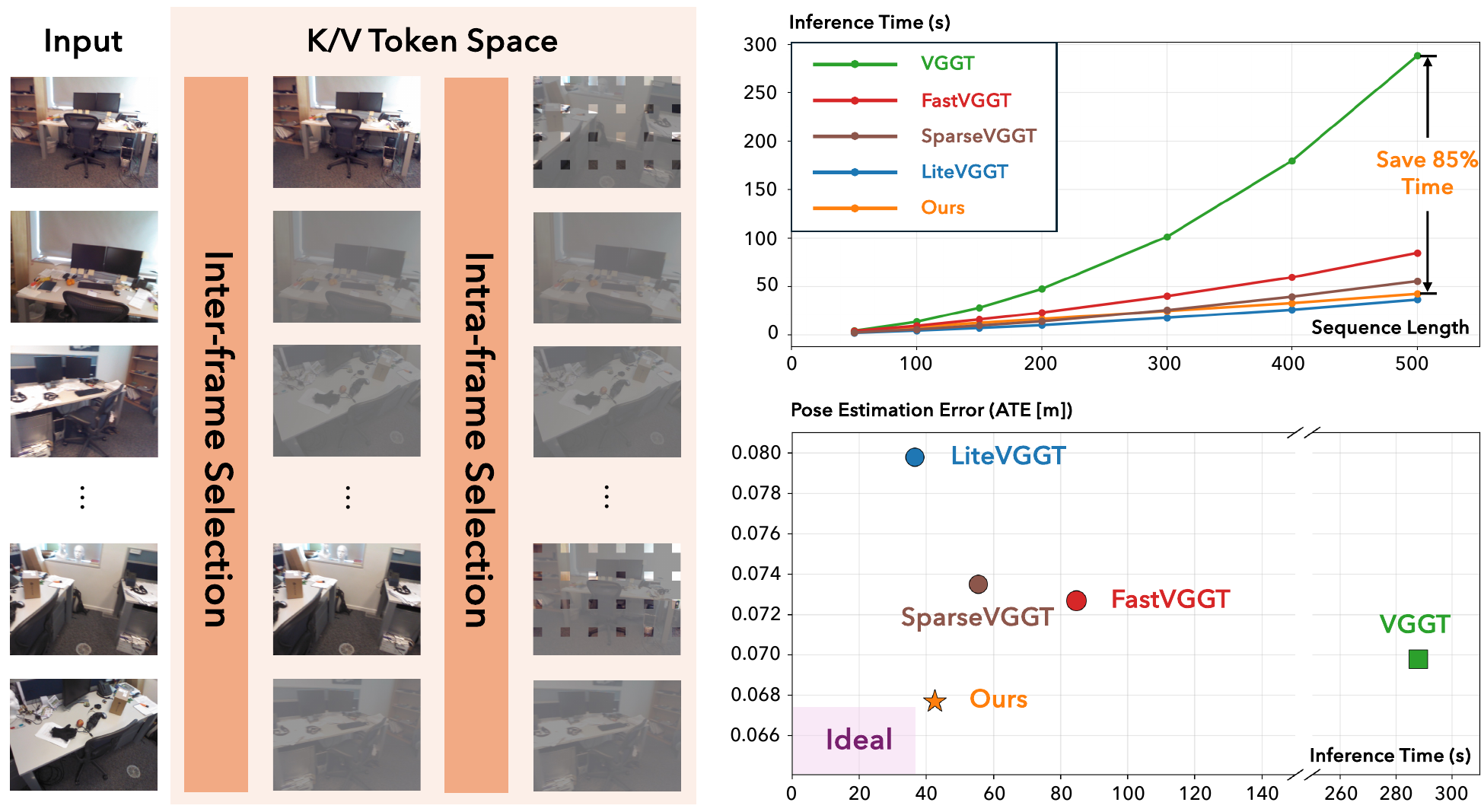}
    \caption{We accelerate visual geometry transformers via a two-stage hierarchical token selection scheme: inter-frame selection followed by intra-frame selection. Our training-free method scales near-linearly with the number of input frames, substantially improving the efficiency of visual geometry transformers with a comparable acceleration ratio with LiteVGGT~\cite{shu2025litevggtboostingvanillavggt}, which requires costly full model training. Overall, our method achieves a superior trade-off between efficiency and accuracy.}
    \label{fig:teaser}
\end{figure*}

Visual geometry transformers~\cite{wang2025vggtvisualgeometrygrounded, wang2025pi3permutationequivariantvisualgeometry, lin2026depth, keetha2025mapanything} are models capable of predicting key 3D attributes (\eg, camera parameters, point maps, depth maps) from multiple views of a scene in a single forward pass. Although these models serve as substantially faster solutions than previous alternatives~\cite{colmap}, they still suffer from prohibitively long inference time when increasing the number of processed frames. This limitation stems from the global attention layers inside these models. While these global attention layers enable effective information aggregation across views, they also exhibit quadratic computational complexity $\mathcal{O}(N^2 L^2)$ in the number of input frames $N$ and per-frame tokens $L$. As a result, global attention becomes the dominant bottleneck, causing inference cost to grow rapidly with the number of input images, and ultimately constraining the efficiency of visual geometry transformers, as illustrated in \Cref{fig:teaser}.

To address this challenge in a principled and generalizable manner, we formulate our problem as follows: in the global attention layers of visual geometry transformers, \emph{given a limited budget of key/value tokens with which each query can interact, how should these tokens be selected?} Our study, Good Token Hunting \textit{(\methodname)}, investigates this question by exploring and analyzing various token selection strategies. Existing solutions~\cite{shen2025fastvggttrainingfreeaccelerationvisual} directly select tokens from the full set across all frames and require computationally heavy inspection of all tokens. In contrast, we leverage a two-stage hierarchical token selection scheme. The first stage performs \textit{inter-frame selection} at the frame level, determining key/value tokens from which frames should be retained. This is a non-trivial task because several intuitive strategies, including similarity-based or activation-based criteria, incur significant performance degradation. Instead, inspired by keyframe-based SLAM~\cite{keyframe_slam_1, keyframe_slam_2} systems, we propose selecting a collection of frames that are as diverse as possible to ensure broad scene coverage. Empirically, this diversity-driven approach proves to be an effective inter-frame selection strategy under tight token budgets, largely preserving the performance from base models while significantly reducing computational cost.

After completing token selection on the frame level, we perform \textit{intra-frame selection} to further improve efficiency by discarding more key/value tokens within each selected frame. We first discover that uniformly downsampling across all global attention layers induces a non-negligible performance drop. To mitigate this issue, we conduct an analysis on the global attention patterns within each layer. We find that early layers exhibit heavily diluted attention, which is a phenomenon also found in language models~\cite{yao2023selfattentionnetworksprocessbounded, zhang2024selectiveattentionenhancingtransformer, chi-etal-2024-attention,kong2025token}, while middle and late layers tend to display spiking values in the attention map. These observations motivate a layer-adaptive intra-frame selection strategy, in which different levels of token pruning are applied across different layers. In particular, in layers with highly activated tokens, we adopt more conservative strategies to avoid discarding important tokens before the actual attention scores are calculated.
Combined with the preceding inter-frame selection stage, this two-stage hierarchical design, as demonstrated in \Cref{fig:teaser}, substantially improves efficiency of visual geometry transformers. For example, on scenes with 500 input frames, our method reduces the inference time of the base VGGT~\cite{wang2025vggtvisualgeometrygrounded} model by over 85\%, while achieving a more favorable trade-off between inference speed and performance compared to existing acceleration approaches~\cite{wang2025sparsevggt, shen2025fastvggttrainingfreeaccelerationvisual, shu2025litevggtboostingvanillavggt}.

In summary, our work makes four contributions. \textbf{(1)} First, we cast the speedup of visual geometry transformers into a straightforward yet general formulation by constraining the number of key/value tokens each query interacts with in global attention layers.
\textbf{(2)} Second, to solve this problem, we introduce a novel hierarchical token selection strategy consisting of inter-frame and intra-frame selection for global attention layers. \textbf{(3)} Third, we provide a systematic exploration of token selection strategies, showing that diversity-based solutions are well-suited for inter-frame selection, while layer-adaptive strategies with different levels of token pruning is critical for intra-frame selection. These findings offer practical guidance for improving both efficiency and performance of visual geometry transformers. \textbf{(4)} Finally, comprehensive experimental results demonstrate that our training-free \methodname solution achieves superior trade-off between efficiency and performance for accelerating visual geometry transformers compared to existing methods, delivering competitive inference speed improvement with minimal performance compromise.

%% file: secs/2_related.tex
\section{Related works}
\label{sec:related}

\textbf{Feed-forward 3D reconstruction.} Multi-view 3D reconstruction tasks, like Structure-from-Motion (SfM) and Multi-view Stereo (MVS), are traditionally solved using complex pipelines involving optimization~\cite{colmap}. While
these methods achieve high accuracy under favorable conditions, they rely on iterative non-linear optimization steps like bundle adjustment~\cite{bundle_adjustment}.
Recent emergence of feed-forward 3D reconstruction models mark a fundamental departure from solving for geometry through optimization. DUSt3R~\cite{wang2024dust3rgeometric3dvision} and its follow-up works~\cite{mast3r_eccv24, lu2024align3r, Tang_2025_CVPR, Jang_2025_CVPR, mast3r-sfm, zhang2025flarefeedforwardgeometryappearance, test3r_neurips, must3r_cvpr25} pioneered this paradigm by predicting pairwise 3D point maps from image pairs using neural networks~\cite{weinzaepfel2023crocov2improvedcrossview, weinzaepfel2023crocoselfsupervisedpretraining3d}. More recently, \emph{Visual Geometry Transformers} such as VGGT~\cite{wang2025vggtvisualgeometrygrounded} further broadened this paradigm to jointly predict key 3D attributes like cameras, depth, or point maps from multiple images. This formulation inspired subsequent works~\cite{wang2026moe3dmixtureofexpertsmodule3d, wang2025amb3r, peng2025omnivggtomnimodalitydrivenvisual, han2025emergentoutlierviewrejection, gao2025more3dvisualgeometry}, including $\pi^3$~\cite{wang2025pi3permutationequivariantvisualgeometry}, MapAnything~\cite{keetha2025mapanything}, and Depth Anything 3~\cite{lin2026depth}, which explore alternative architectural design choices. This line of work has already been applied in multiple areas~\cite{mahdavian2026uniscaleunifiedscaleaware3d,wu2026mvggtmultimodalvisualgeometry, li2025iggt, liu2025vggtxvggtmeetsdense, alumootil2025dept3rjointdensepoint, chen2026stereovggttrainingfreevisualgeometry, fang2026densr, RnG, yuan2026vggt360geometryconsistentzeroshotpanoramic}. 
For example, streaming reconstruction methods~\cite{khafizov2025gcut3rguided3dreconstruction,yuan2026infinitevggtvisualgeometrygrounded,lu2026ovggto1constantcoststreaming,xu2026framevggtframeevidencerolling,fang2026incvggt,liu2026streamcachevggtstreamingvisualgeometry,wang2026stac,gelencsérhorváth2026scenevggtvggtbasedonline3d} are realized through causal and autoregressive models~\cite{lan2025stream3rscalablesequential3d,streaming,yan2026omnistreammasteringperceptionreconstruction,cheng2026longstreamlongsequencestreamingautoregressive}, preserved geometric context~\cite{li2025wint3rwindowbasedstreamingreconstruction,chen2026geometriccontexttransformerstreaming,qiu2026slarmstreaminglanguagealignedreconstruction}, recurrent memory~\cite{long3r_iccv,liu2026mem3rstreaming3dreconstruction}, and adaptive state updates~\cite{chen2025ttt3r3dreconstructiontesttime,jin2026filt3rlatentstateadaptive,dong2026memixwritinglessremembering,zheng2026ttsa3rtrainingfreetemporalspatialadaptive}. Dynamic scene reconstruction methods~\cite{jiang2025geo4d, wang20254dvggtgeneralfoundationmodel,wu20254dlangvggt4dlanguagevisualgeometry,sun2026densedynamicscenereconstruction,xu2024das3rdynamicsawaregaussiansplatting} leverage attention-derived motion cues~\cite{easi3r,shen2025mut3rmotionawareupdatingtransformer,hu2025vggt4dminingmotioncues,xiong2026vggtmotionmotionawarecalibrationfreemonocular,ding2025laserlayerwisescalealignment}, dynamic-aware aggregation and uncertainty modeling~\cite{zhou2025page4ddisentangledposegeometry,fang2026moremotionawarefeedforward4d,zang2026robust4dvisualgeometry}, dynamic point maps and temporal matching~\cite{sucar2025dynamicpointmapsversatile,POMATO_Zhang_2025_ICCV,he2026dynamicvggtlearningdynamicpoint}, and explicit scene flow and motion representations~\cite{qian2026flow4runifying4dreconstruction,karhade2025any4dunifiedfeedforwardmetric}.
Besides, there are also works focusing on human-centric reconstruction~\cite{chen2025human3r, zhang2025odhsronlinedense3d}, autonomous driving~\cite{jia2025drivevggtvisualgeometrytransformer, yu2026recondrivefastfeedforward4d, zuo2025dvgt}, visual relocalization~\cite{xu2026gpavggtadaptingvggtlargescale, deng2025relocvggtvisualrelocalizationgeometry, miao2026trajvg3dtrajectorycoupledvisual, dinya2025buildingtemporallycoherent3d}, and odometry~\cite{dai2026keyframebasedfeedforwardvisualodometry, pan2026hyvggtvotightlycoupledhybrid}. This breadth of applications demonstrates the growing importance and versatility of visual geometry transformers. The substantial computational cost when using a large number of input images is one of the main obstacles towards even broader impact for these models.

\textbf{Efficiency improvement on visual geometry transformers.}
To address this challenge, a growing body of research~\cite{Yang_2025_Fast3R, wang2025httmheadwisetemporaltoken, mahdi2025evict3rtrainingfreetokeneviction, lee2025swiftvggtscalablevisualgeometry, kim2026hessheadsensitivityscore, li2025analyzingmechanismattentioncollapse, shu2025litevggtboostingvanillavggt} aims at improving efficiency to make visual geometry transformers practical at scale. For example, FastVGGT~\cite{shen2025fastvggttrainingfreeaccelerationvisual} introduced a training-free token merging scheme that preserves reference and salient tokens while merging the rest. Other approaches such as SparseVGGT~\cite{sun2025avggtrethinkingglobalattention} inspect the behavior of global attention and introduce specific attention calculation and token pruning mechanisms to speed up inference. AVGGT~\cite{sun2025avggtrethinkingglobalattention} replaces early global attention with frame attention and subsamples key/value tokens with diagonal preservation and mean filling. Compression-based approaches reduce the inference cost through low-bit compression, including quantized VGGT~\cite{feng2025quantizedvisualgeometrygrounded} and tail-aware quantization~\cite{pan2026tailawareposttrainingquantization3d}. In contrast, methods~\cite{wang2025flashvggtefficientscalablevisual} like LiteVGGT~\cite{shu2025litevggtboostingvanillavggt} and Speed3R~\cite{ren2026speed3r} improve efficiency by retraining the model with additional priors or architectural constraints, so that the global attention layers operate on fewer tokens. Unlike these works, we adopt a training-free approach that allows for selecting a few key/value tokens within a limited budget that each query can interact with.

%% file: secs/3_method_arxiv.tex
\section{\methodname: Token selection for global attention}
\label{sec:method}

\begin{figure*}[t]
    \centering
    \includegraphics[width=\linewidth]{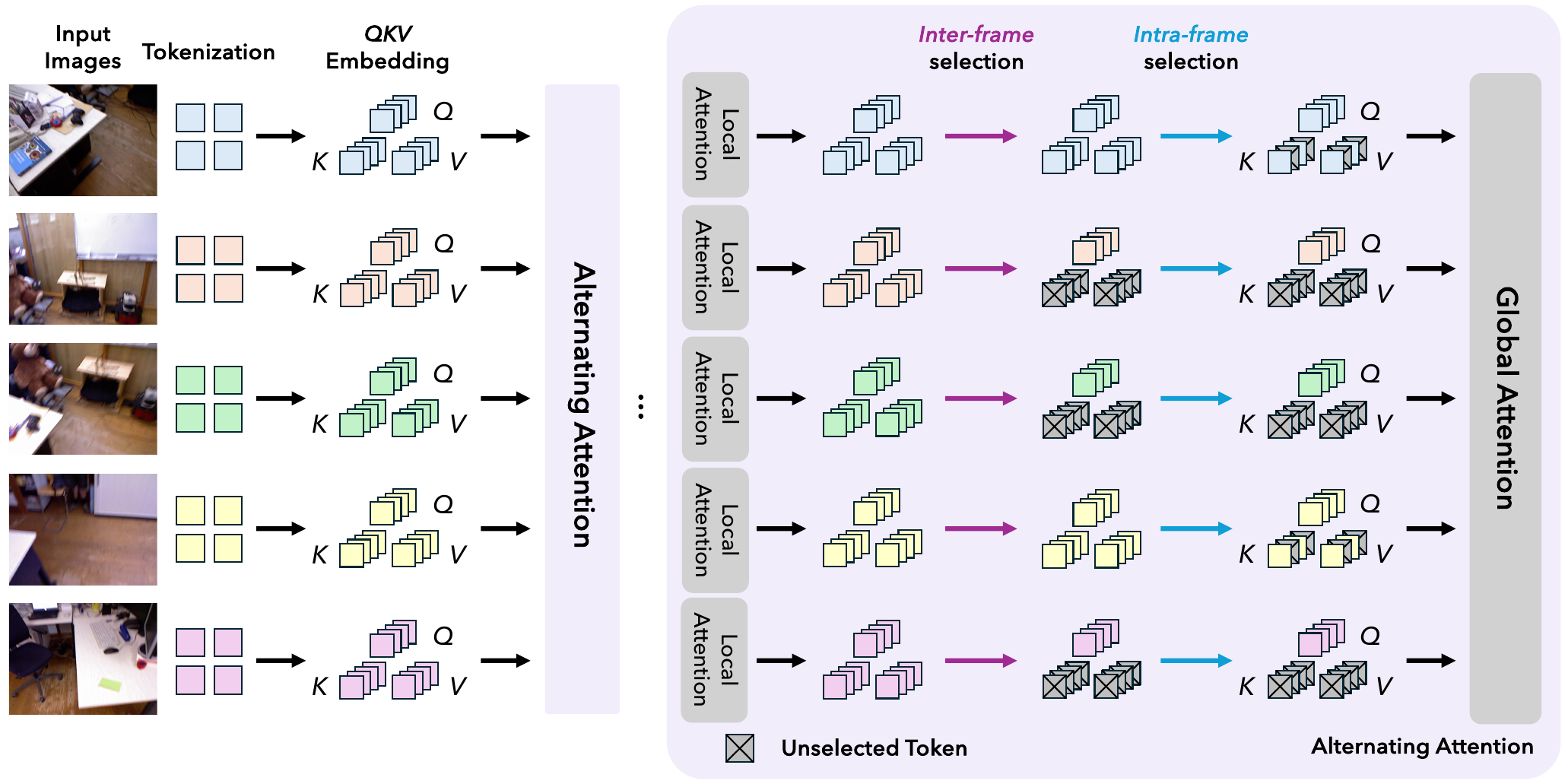}
    \caption{Pipeline of \methodname. Token selection is performed in the \textit{K/V} space prior to the global attention layers, to determine which key/value tokens each query token interacts with. Our approach follows a two-stage hierarchical design: \textit{inter-frame selection} first conducts frame-level selection, while \textit{intra-frame selection} subsequently discard more tokens within each selected frame. 
    }
    \label{fig:pipeline}
\end{figure*}

\subsection{Preliminaries and problem formulation}
\label{sec:prelim}

\textbf{Visual geometry transformers} take in $N$ images $\mathcal{I}=\left\{I_n\right\}_{i=1}^N$ capturing a scene as input, and predicts geometric properties for each frame such as camera pose $[\mathbf{R}_i|\mathbf{t}_i]$, point maps
$\mathbf{P}_{i}$, \etc, depending on the model design. Specifically, each image is first patchified
into $L$ spatial tokens, optionally concatenated with special tokens (\eg, camera tokens). These tokens are then processed by a stack of frame-wise attention layers, which operate independently within each frame, and global attention layers, which jointly operate across all tokens from all frames. After cross-view information aggregation, dedicated task-specific heads decode each geometric property from the processed representations.

\textbf{Computational bottleneck.} As also illustrated in prior work~\cite{shen2025fastvggttrainingfreeaccelerationvisual, shu2025litevggtboostingvanillavggt}, the inference efficiency of visual geometry transformers is primarily constrained by the global attention layers, which compute attention over the entire set of $N\times L$ tokens ($N$ is the number of input frames, and $L$ is the number of tokens per frame), along with any additional special tokens. This results in a quadratic computational complexity of $\mathcal{O}(N^2 L^2)$, which is the central bottleneck addressed in this work.

\textbf{Problem formulation.} To address this challenge in a general and principled manner, we adopt the following simple formulation: restricting the number of key/value tokens that each query attends to within each global attention layer. Rather than directly selecting tokens from the entire set across all frames, which is inefficient and suboptimal as it requires computationally heavy scan on all tokens, we employ a hierarchical selection strategy. We first perform inter-frame selection (\Cref{sec:interframe}) to select a set of frames. Then, we apply intra-frame selection (\Cref{sec:intraframe}) within each selected frame to further discard more tokens. This two-stage design enables efficient token selection under the budget constraint while preserving essential information.


\textbf{Preliminary experiment setting.} In Sections~\ref{sec:interframe} and~\ref{sec:intraframe}, we conduct preliminary experiments to systematically analyze inter-frame and intra-frame token selection strategies. We evaluate camera pose estimation on the 7-Scenes~\cite{7scenes} dataset. We sample every 2 frames of the image sequences, resulting in 500 frames per scene (with the exception of two scenes containing 250 frames). We follow previous works~\cite{wang2025pi3permutationequivariantvisualgeometry, shen2025fastvggttrainingfreeaccelerationvisual} and adopt the metrics of Absolute Trajectory Error (ATE), Relative Pose Error in rotation (RPE-rot) and translation (RPE-trans).

\subsection{Inter-frame selection: Hunting for good frames}
\label{sec:interframe}

\begin{figure}[t]
\centering
\begin{minipage}[t]{0.55\textwidth}
    \rowcolors{2}{white}{uoftcoolgray!25}
    \centering
    \vspace{11pt}
    \captionof{table}{Comparison of intuitive inter-frame selection strategies on 7-Scenes~\cite{7scenes} for camera pose estimation. The budget restricts each query to attend to key/value tokens from $K=25$ frames.}
    \resizebox{\linewidth}{!}{
        \begin{tabular}{lccc}
        \toprule[0.17em]
        \textbf{Strategy}  & 
        \textbf{ATE ($\downarrow$)} & 
        \textbf{RPE-rot ($\downarrow$)} & 
        \textbf{RPE-trans ($\downarrow$)} \\ 
        \midrule
        \multicolumn{4}{l}{\textbf{(1) Temporal proximity}} \\
        Nearest & 0.7588 & 1.8485 & 0.0563 \\
        \midrule
        \multicolumn{4}{l}{\textbf{(2) Co-visibility-based selection}} \\
        \textbf{(2a)} High co-visibility & 0.3813 & 2.9934 & 0.1197 \\
        \textbf{(2b)} Low co-visibility & 0.1840 & 2.4761 & 0.1038\\
        \midrule
        \multicolumn{4}{l}{\textbf{(3) Attention-based selection}} \\
        \textbf{(3a)} Max pooling & 0.3879 & 7.2494 & 0.1257\\
        \textbf{(3b)} Mean pooling  & 0.3627 & 7.2988 & 0.0988\\
        \midrule
        \multicolumn{4}{l}{\textbf{(4) Selection strategy from existing work} \textit{\textbf{(ordered input)}}} \\
        VGGT-SLAM & 0.0741 & 0.5958 & 0.0203\\
        \midrule
        Diversity-based (Ours) & \textbf{0.0676} & \textbf{0.4421} & \textbf{0.0167}\\
        \midrule
        VGGT \textit{(Base Model)}& \textcolor{gray}{0.0698} & \textcolor{gray}{0.4953} & \textcolor{gray}{0.0178}\\
        \bottomrule[0.17em]
        \end{tabular}
    }
    \label{tab:intuitive_solutions}
\end{minipage}
\hfill
\begin{minipage}[t]{0.43\textwidth}
\vspace{1pt}
    \centering
    \includegraphics[width =0.8\linewidth]{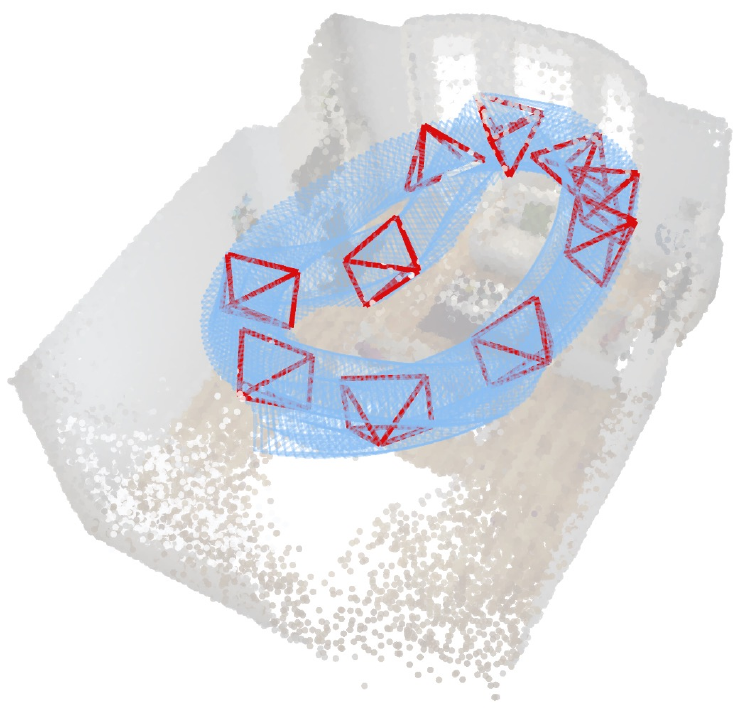}
    \caption{
    Illustration of inter-frame selection with $K=10$: the selected views (\textcolor{uoftwarmred}{red}) form a diverse subset of the full set of views (\textcolor{uoftsecondaryblue}{blue}), maximizing view-space coverage under a limited budget.}
\vspace{-10pt}
    \label{fig:inter_frame_selection}
\end{minipage}

\end{figure}

\textbf{Intuitive strategies.} The first stage in token selection is inter-frame selection, to determine what frames to keep for further processing. First, we evaluate several intuitive strategies: \textbf{(1)} selecting temporally adjacent frames (only applicable to ordered sequences); \textbf{(2)} selecting frames based on co-visibility, with variants of \textbf{(2a)} selecting frames that are \textit{most} co-visible with the current frame; \textbf{(2b)} selecting frames that are \textit{least} co-visible with the current frame; \textbf{(3)} selecting frames based on attention activation, split into \textbf{(3a)} selecting based on the \textit{maximum} attention score and \textbf{(3b)} selecting based on the \textit{mean} attention score; and \textbf{(4)} selecting frames following existing work VGGT-SLAM~\cite{maggio2025vggtslam}. Specifically, as VGGT-SLAM requires sequentially ordered input frames, it is evaluated on ordered sequences while the other settings do not necessarily require. For co-visibility approximation, we utilize the place recognition model~\cite{Berton_2025_MegaLoc} to extract features for each input image. The similarity between features serves as the proxy for the frame overlap, indicating the co-visibility between image pairs. For this preliminary analysis, we set the budget of selected frames to be $K=25$ from the 7-Scenes sequences of 250/500 frames, meaning that we allow each query to only interact with key/value tokens from 25 frames in the global attention layers. At this level of sparsification, maintaining decent performance after frame selection is non-trivial. As reported in~\Cref{tab:intuitive_solutions}, the intuitive strategies from \textbf{(1)} to \textbf{(3)} all lead to substantial performance degradation. For \textbf{(4)}, it selects frames that have disparity only with the last selected keyframe, ignoring other already selected frames in the set. In contrast, our strategy optimizes for a diverse set of frames where each frame in the selected set is targeted at capturing a different portion of the scene from any other selected frame. Also, we point out that the purpose of frame selection in our paper is fundamentally different from the VGGT-SLAM paper, as in VGGT-SLAM, keyframe selection is for obtaining submaps that run VGGT independently, after which alignment between submaps are performed. In our approach, frame selection is used to determine key/value tokens from which frames need to be preserved, for accelerating the global attention layers inside VGGT, and all frames are still processed by VGGT in a single pass. Therefore, the metrics reported in~\Cref{tab:intuitive_solutions} differ from the VGGT-SLAM paper.

\textbf{Diversity-based frame selection.} In contrast to the above strategies, our intuition is to select a set of frames, within a given budget, that can maximize view-space coverage. Formally, given $N$ images with $d$-dimensional features $\{f_i\}_{i=1}^N$ extracted by the aforementioned place recognition model, we define the \textit{cosine distance} between two images as
\begin{equation}
d(i, j) \;=\; 1 - \frac{\langle f_i, f_j \rangle}{\|f_i\|_2 \,\|f_j\|_2}.
\label{eq:cos-dist}
\end{equation}

Under a budget that allows each query to attend to tokens from $K$ frames, we seek the subset
$S^\star \subseteq \{1, \dots, N\}$ with $|S^\star| = K$ that minimizes the
largest distance from any frame to its nearest selected frame:
\begin{equation}
S^\star \;=\; \argmin_{{S \subseteq \{1,\dots,N\}, |S| = K}}\;
              \max_{i \in \{1,\dots,N\}}\;
              \min_{j \in S}\; d(i, j).
\label{eq:kcenter}
\end{equation}

Since Equation~\ref{eq:kcenter} is the classical NP-hard ``$K$-center'' objective, we adopt the similar greedy farthest point sampling (FPS) heuristic~\cite{Gonzalez1985ClusteringTM}, widely used in point cloud processing, which iteratively selects the frame farthest from the current selected set. For details, we refer to~\Cref{alg:diverse-frames} in the Appendix.



From the results in~\Cref{tab:intuitive_solutions}, we can observe that our inter-frame selection strategy greatly outperforms the intuitive alternatives. These selected frames serve as ``anchors'', as illustrated in~\Cref{fig:inter_frame_selection}, providing broad view-space coverage of the scene with a set of views within a limited budget. Moreover, these ``anchors'' supporting the whole scene are expected to be consistent across different queries, suggesting that a common set of reference views across all tokens is beneficial for cross-view representation processing within visual geometry transformers.


\subsection{Intra-frame token selection: Preserving necessary tokens}
\label{sec:intraframe}
\begin{wraptable}[11]{r}{0.52\linewidth}
    \vspace{-11pt}
    \centering
    \caption{Experimental results with the uniform intra-frame token selection strategy across all global attention layers following AVGGT~\cite{shu2025litevggtboostingvanillavggt}.}
    \rowcolors{2}{white}{uoftcoolgray!25}
    \resizebox{\linewidth}{!}{
    \begin{tabular}{>{\centering\arraybackslash}p{0.1\textwidth}cccc}
        \toprule[0.17em]
        \textbf{$K$}  & \textbf{$\sigma$}  & 
        \textbf{ATE ($\downarrow$)} & 
        \textbf{RPE-rot ($\downarrow$)} & 
        \textbf{RPE-trans ($\downarrow$)} \\ 
        \midrule
        10 & 2 & 0.1031 & 1.4347 & 0.0295\\
        10 & 3 & 0.2414 & 3.7025 & 0.0564\\
        25 & 2 & 0.0831 & 0.6630  & 0.0213\\
        25 & 3 & 0.1393 & 1.8891 & 0.0434\\
        \midrule
        \multicolumn{2}{l}{VGGT \textit{(Base Model)}}  & \textcolor{gray}{0.0698} & \textcolor{gray}{0.4953} & \textcolor{gray}{0.0178}\\
        \bottomrule[0.17em]
    \end{tabular}
    }\label{tab:aggressive_intra_frame_avggt}
    \vspace{-10pt}
\end{wraptable}
\textbf{Performance drop with intra-frame downsampling.} Having determined which frames to retain, we turn to the second stage of token selection: identifying which tokens within each selected frame can be further discarded. Existing work~\cite{sun2025avggtrethinkingglobalattention} suggests that we can apply intra-frame downsampling within all global attention layers by subsampling token maps. Concretely, tokens are downsampled by a factor of $\sigma$ along both the height and width dimensions, reducing a feature map with the original size $h\times w$ to $\lfloor \frac{h}{\sigma}\times\frac{w}{\sigma}\rfloor$. Following their approach, we perform downsampling across all global attention layers. However, we observe a noticeable performance drop, as reported in \Cref{tab:aggressive_intra_frame_avggt}. Even a modest downsampling factor of $\sigma=2$ leads to measurable performance degradation.

\begin{figure*}[t]
    \centering
    \includegraphics[width=\linewidth]{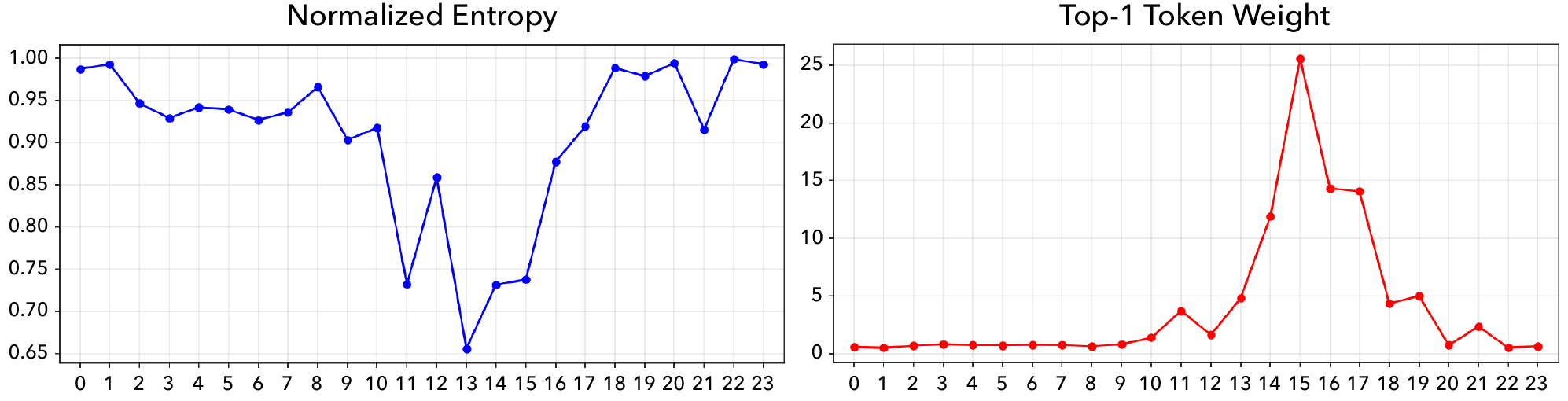}
    \caption{
    Attention pattern analysis of global attention layers (0-23) within VGGT~\cite{wang2025vggtvisualgeometrygrounded}. Early layers show diluted, near-uniform attention distributions, whereas middle layers have spiking values.}
    \label{fig:attention_pattern}
    \vspace{-1em}
\end{figure*}

\textbf{Attention pattern analysis.} To understand the reason behind the performance degradation after intra-frame downsampling, we inspect the attention patterns within the global attention layers. Specifically, we report two statistics in~\Cref{fig:attention_pattern}: \textit{normalized entropy} $\mathcal{H}_{\rm norm}\in[0, 1]$ and \textit{top-1 token weight}, computed over a set of sampled query tokens and attention heads for each layer. The \textit{normalized entropy} is formalized as 
\begin{equation}
    \mathcal{H}_{\rm norm} = \frac{\sum_{0\leqslant h<H,0\leqslant q<Q} \mathcal{H}(h,q)}{H\cdot Q \cdot \mathcal{H}_{\max}},
\end{equation}
where $\mathcal{H}_{\max}=\log{(NL)}$ is the maximum possible entropy over all key tokens, with $N$ being the number of frames and $L$ the number of tokens per frame. $\mathcal{H}(h,q)$ represents the entropy of the attention scores on attention head $h$ and query $q$. $H=4$ and $Q=50$ are the number of sampled attention heads and query tokens for calculating these statistics.


\begin{wraptable}[14]{r}{0.55\linewidth}
    \vspace{-10pt}
    \centering
    \caption{Performance analysis on different intra-frame strategies applied on different sets of VGGT~\cite{wang2025vggtvisualgeometrygrounded} layers, with $K=25$.}
    \rowcolors{2}{white}{uoftcoolgray!25}
    \resizebox{\linewidth}{!}{
    \begin{tabular}{cccccc}
        \toprule[0.17em]
        \textbf{$\sigma$} & \textbf{Strategy}  & \textbf{Layers} &
        \textbf{ATE ($\downarrow$)} & 
        \textbf{RPE rot ($\downarrow$)} & 
        \textbf{RPE trans ($\downarrow$)} \\ 
        \midrule
        2 & \textit{Standard}& 0-8 & 0.0676 & 0.4427 & 0.0168\\
        2 & \textit{Standard} & 9-16 &  0.0792 & 0.9539 & 0.0239\\
        2 & \textit{Activation} & 9-16 & 0.0687  & 0.4664 & 0.0172\\
        2 & \textit{Standard}& 17-23 & 0.0715  & 0.4463 & 0.0168\\
        3 & \textit{Standard}&0-8  & 0.0679 & 0.4486 & 0.0168\\
        3 & \textit{Standard} &9-16 & 0.1234  &1.6722  & 0.0416\\
        3 & \textit{Activation} & 9-16 &  0.0711 &  0.9163 & 0.0207\\
        3 & \textit{Standard}& 17-23 &  0.0743 & 0.4527 & 0.0172\\
        \midrule
        \multicolumn{3}{l}{VGGT~\cite{wang2025vggtvisualgeometrygrounded} \textit{(Base Model)}} & \textcolor{gray}{0.0698} & \textcolor{gray}{0.4953} & \textcolor{gray}{0.0178}\\
        \bottomrule[0.17em]
        \end{tabular}}\label{tab:different_layer_intra_frame_avggt}
    \vspace{-10pt}
\end{wraptable}
As shown in~\Cref{fig:attention_pattern}, early global attention layers exhibit a diluted, near-uniform attention pattern, whereas middle and later layers showcase a sharp attention pattern with spiking attention values. This observation suggests that, if the token downsampling in the middle and late layers discard tokens that are highly activated, their attention pattern will be severely disrupted, resulting in a performance degradation. This hypothesis is also supported by the comparison between the \textit{Standard} and \textit{Activation} strategies in~\Cref{tab:different_layer_intra_frame_avggt}, where the \textit{Activation} preserves the same fraction of tokens ($\frac{1}{4}$ for $\sigma=2$, $\frac{1}{9}$ for $\sigma=3$) by selecting tokens with the highest attention activations, while \textit{Standard} uniformly drops tokens in both height and width dimensions. The substantially reduced performance compromise of \textit{Activation} for the middle layers indicates that token selection in layers with spiking attention values needs to be carefully designed. However, since identifying highly activated tokens requires computing attention scores in advance, which is time-consuming, \textit{Activation} can only serve as a validation for our hypothesis, instead of a practical and efficient solution. In contrast, since the attention is diluted in the early layers without highly activated tokens, more aggressive intra-frame downsampling can be safely applied in these layers while still largely preserving the performance, as supported by the results in~\Cref{tab:different_layer_intra_frame_avggt}.



\textbf{Layer-adaptive intra-frame strategy.} The attention patterns in~\Cref{fig:attention_pattern} reveal that early layers of visual geometry transformers tend to have diluted attention patterns, where we can safely perform intra-frame downsampling without concerning about dropping highly activated tokens. Furthermore, we observe that the very first few layers have normalized entropy values close to 1, indicating that global attention in these layers can barely function for cross-view interaction. Following~\cite{sun2025avggtrethinkingglobalattention}, we can replace these global attention layers with local attention operating within each frame to further save compute. Therefore, to formalize this design, we introduce two thresholds, $l_{\rm local}$ and $l_{\rm sample}$, to determine the intra-frame strategies applied to each layer. For layers with index $l< l_{\rm local}$, we replace global attention with local attention, which is the more aggressive intra-frame strategy to speed up the inference. For layers with index $l_{\rm local}\leqslant l< l_{\rm sample}$, we apply intra-frame downsampling with a selected factor. This layer-adaptive strategy balances efficiency and accuracy by aligning the levels of token pruning with the underlying attention characteristics for different global attention layers.


%% file: secs/4_comparison.tex
\section{Experiments}
\label{sec:experiments}

\subsection{Experimental setup}

\textbf{Implementation details.} We choose two representative visual geometry transformers VGGT~\cite{wang2025vggtvisualgeometrygrounded} and $\pi^3$~\cite{wang2025pi3permutationequivariantvisualgeometry} as base models for evaluation. For comparisons with other methods in \Cref{sec:compare_with_others}, we choose $K=25$ and $\sigma\in\{2,3\}$ for a relatively fixed budget of selected tokens. In the analysis in \Cref{sec:ablation}, we further show the model performance under different budgets. Unless otherwise specified, we set the layer thresholds to $l_{\rm local}=2$ and $l_{\rm sample}=9$, but also demonstrate in \Cref{sec:ablation} that the performance is robust to these thresholds. All experiments are conducted on a single NVIDIA L40S GPU with 48GB CUDA memory.


\textbf{Tasks, metrics, and datasets.} Beyond the camera pose estimation task already introduced in \Cref{sec:prelim}, we also evaluate our method on 3D point cloud reconstruction and video depth estimation. Following previous works~\cite{wang2025pi3permutationequivariantvisualgeometry}, we adopt the mean and median values of Accuracy (Acc), Completion (Comp), and Normal Consistency (NC) as evaluation metrics for 3D reconstruction, and Absolute Relative Error (Abs Rel), Root Mean Squared Error (RMSE), Log RMSE, Squared Relative Error (Sq Rel), and prediction accuracy at the threshold of $\delta < 1.25$ for video depth estimation. Detailed explanations on the metrics of all three tasks can be referred in~\Cref{sec:evaluation_metrics} in the Appendix. Experiments are conducted on a diverse set of benchmarks, including 7-Scenes~\cite{7scenes}, Neural RGB-D~\cite{Azinovic_2022_CVPR}, TUM-Dynamics~\cite{tum_dynamics}, and Bonn~\cite{bonn}.


\textbf{Baseline methods.} We compare against the state-of-the-arts for accelerating visual geometry transformers, including FastVGGT~\cite{shen2025fastvggttrainingfreeaccelerationvisual}, SparseVGGT~\cite{wang2025sparsevggt}, Co-Me~\cite{chen2025comeconfidenceguidedtokenmerging}, LiteVGGT~\cite{shu2025litevggtboostingvanillavggt}, and Speed3R~\cite{ren2026speed3r}. We follow the default sparsification settings adopted in these methods. For SparseVGGT, we report results with sparsity ratio (SR) of 50\% and 75\% using a CDF threshold of 0.9. Among these methods, LiteVGGT and Speed3R require full model retraining, typically taking several days on a multi-GPU setup (8 high-performance GPUs). Co-Me involves lightweight training under 1 hour on an NVIDIA A100 GPU. In contrast, FastVGGT, SparseVGGT, and our method are training-free. SparseVGGT can also be applied to $\pi^3$ (denoted as Sparse-$\pi^3$), whereas Speed3R is only available with the $\pi^3$ checkpoint.

\begin{table}[t]
    \vspace{-1em}
    \rowcolors{2}{white}{uoftcoolgray!25}
    \centering
    \caption{Quantitative comparisons on camera pose estimation together with the 500-frame sequence inference time on a single NVIDIA L40S GPU. Best is \textbf{bold} and the second best is \underline{underlined}, excluding the base model row.} 
    \vspace{5pt}
    \resizebox{\linewidth}{!}{
    \begin{tabular}{lcccccccccc}
        \toprule[0.17em]
        \cellcolor{white} &
        \multicolumn{3}{c}{\textbf{7-Scenes}} &
        \multicolumn{3}{c}{\textbf{Neural RGB-D}} &
        \multicolumn{3}{c}{\textbf{TUM-Dynamics}} &  \multicolumn{1}{c}{\textbf{Efficiency}}\\
        \cmidrule(r){2-4} \cmidrule(r){5-7} \cmidrule(r){8-10} \cmidrule(r){11-11}
        \cellcolor{white}\multirow{-2}{*}[0.5ex]{\textbf{Method}} &
        \textbf{ATE ($\downarrow$)} & \textbf{RPE-rot ($\downarrow$)} & \textbf{RPE-trans ($\downarrow$)} &
        \textbf{ATE ($\downarrow$)} & \textbf{RPE-rot ($\downarrow$)} & \textbf{RPE-trans ($\downarrow$)} &
        \textbf{ATE ($\downarrow$)} & \textbf{RPE-rot ($\downarrow$)} & \textbf{RPE-trans ($\downarrow$)} &  \makecell{\textbf{500-frame}\\\textbf{Inference Time ($\downarrow$)}} \\
        \midrule
        VGGT~\cite{wang2025vggtvisualgeometrygrounded} \textit{(Base Model)}  & \textcolor{gray}{0.0698}& \textcolor{gray}{0.4953}& \textcolor{gray}{0.0178}& \textcolor{gray}{0.0374}& \textcolor{gray}{0.2934}& \textcolor{gray}{0.0186}& \textcolor{gray}{0.0118}& \textcolor{gray}{0.3083}& \textcolor{gray}{0.0098} & \textcolor{gray}{288.0s}\\
        \midrule
        FastVGGT~\cite{shen2025fastvggttrainingfreeaccelerationvisual} &0.0727 & \textbf{0.4254}& \textbf{0.0159}& 0.0377& \underline{0.1985}& \underline{0.0168}& 0.0127 & 0.3154 & 0.0108 & 84.6s\\
        SparseVGGT~\cite{wang2025sparsevggt} (SR: 50\%)& 0.0723& 0.4608& 0.0167& 0.0402& 0.2946& 0.0202 &0.0125 & 0.3114& \underline{0.0102} & 80.3s\\
        SparseVGGT~\cite{wang2025sparsevggt} (SR: 75\%)& 0.0735& 0.4583& 0.0169& 0.0462& 0.2717& 0.0192 & 0.0127& 0.3120& 0.0103 & 55.4s\\
        Co-Me~\cite{chen2025comeconfidenceguidedtokenmerging} &0.0870 & 0.8105& 0.0340& 0.0626& 0.4567& 0.0336& 0.0156& 0.3438& 0.0146 & 84.2s\\
        LiteVGGT~\cite{shu2025litevggtboostingvanillavggt} & 0.0798& 0.6888&0.0238 & 0.0531& 0.3311& 0.0247& 0.0145 & 0.3250& 0.0119 & \textbf{36.5s}\\
        \textbf{\methodname (Ours) ($\sigma=2$)} & \textbf{0.0673} & \underline{0.4471} & \underline{0.0165} & \textbf{0.0267}& \textbf{0.1794}& \textbf{0.0162}& \textbf{0.0115}& \underline{0.3087} & \textbf{0.0101} & 44.6s\\
        \textbf{\methodname (Ours) ($\sigma=3$)} & \underline{0.0677}& 0.4495& 0.0166& \underline{0.0270} &0.2409 &0.0176 &\underline{0.0119} & \textbf{0.3075}& \underline{0.0102} & \underline{41.2s}\\
        \midrule
        $\pi^3$~\cite{wang2025pi3permutationequivariantvisualgeometry} \textit{(Base Model)} & \textcolor{gray}{0.0573} &\textcolor{gray}{0.3389} &\textcolor{gray}{0.0105} & \textcolor{gray}{0.0251}& \textcolor{gray}{0.1031}& \textcolor{gray}{0.0098}& \textcolor{gray}{0.0140}& \textcolor{gray}{0.3073}& \textcolor{gray}{0.0088} & \textcolor{gray}{106.5s}\\
        \midrule
        Sparse-$\pi^3$~\cite{wang2025sparsevggt} (SR: 50\%) & 0.0580& \textbf{0.3369}&\textbf{0.0106} & 0.0313& \textbf{0.1182}& \textbf{0.0115}& \textbf{0.0140} &\textbf{0.3068} &0.0090 & 76.4s\\
        Sparse-$\pi^3$~\cite{wang2025sparsevggt} (SR: 75\%) & 0.0594 & \underline{0.3387}& \underline{0.0108}& 0.0478& 0.1250& 0.0124&  \underline{0.0141}& 0.3094&0.0092 & 74.1s\\
        Speed3R~\cite{ren2026speed3r} & 0.0591& 0.3800& 0.0133& 0.0391& 0.1735& 0.0145& 0.0193& 0.3152& 0.0103 & \textbf{17.9s}\\
        \textbf{\methodname (Ours) ($\sigma=2$)} & \underline{0.0579}& 0.3445& 0.0113&\textbf{0.0292} &\underline{0.1190} &\underline{0.0123} & 0.0142& \underline{0.3075}& \textbf{0.0089} & 26.4s\\
        \textbf{\methodname (Ours) ($\sigma=3$)} & \textbf{0.0570}& 0.3428& 0.0112&\textbf{0.0292} &0.1192 &\underline{0.0123} &0.0144 & 0.3083& \textbf{0.0089} & \underline{26.1s}\\
        \bottomrule[0.17em]
    \end{tabular}
    }
    \label{tab:relpose}
\end{table}

\begin{table}[t]
    \rowcolors{4}{white}{uoftcoolgray!25}
    \centering
    \caption{
        Quantitative comparisons on point map estimation. Best is \textbf{bold} and the second best is \underline{underlined}, excluding the base model row.
    }
    \vspace{5pt}
    \resizebox{\columnwidth}{!}{
    \begin{tabular}{lcccccccccccc}
        \toprule[0.17em]
        \cellcolor{white} &
        \multicolumn{6}{c}{\textbf{7-Scenes}} &
        \multicolumn{6}{c}{\textbf{Neural RGB-D}} \\
        \cmidrule(r){2-7} \cmidrule(r){8-13}
        
        \cellcolor{white} &
        \multicolumn{2}{c}{\textbf{Acc ($\downarrow$)}}  &
        \multicolumn{2}{c}{\textbf{Comp ($\downarrow$)}} &
        \multicolumn{2}{c}{\textbf{NC ($\uparrow$)}}     &
        \multicolumn{2}{c}{\textbf{Acc ($\downarrow$)}}  &
        \multicolumn{2}{c}{\textbf{Comp ($\downarrow$)}} &
        \multicolumn{2}{c}{\textbf{NC ($\uparrow$)}}     \\
        \cmidrule(r){2-3} \cmidrule(r){4-5} \cmidrule(r){6-7}
        \cmidrule(r){8-9} \cmidrule(r){10-11} \cmidrule(r){12-13}
        
        \cellcolor{white}\multirow{-3}{*}[1ex]{\textbf{Method}} &
        \textbf{Mean} & \textbf{Med.} &
        \textbf{Mean} & \textbf{Med.} &
        \textbf{Mean} & \textbf{Med.} &
        \textbf{Mean} & \textbf{Med.} &
        \textbf{Mean} & \textbf{Med.} &
        \textbf{Mean} & \textbf{Med.} \\
        \midrule[0.08em]
        VGGT~\cite{wang2025vggtvisualgeometrygrounded} \textit{(Base Model)} & \textcolor{gray}{0.0171} & \textcolor{gray}{0.0038}& \textcolor{gray}{0.0184}& \textcolor{gray}{0.0043}& \textcolor{gray}{0.5568}& \textcolor{gray}{0.5851}&\textcolor{gray}{0.0160} & \textcolor{gray}{0.0099}&\textcolor{gray}{0.0112} & \textcolor{gray}{0.0028}& \textcolor{gray}{0.7508}& \textcolor{gray}{0.8917}\\
        \midrule
        FastVGGT~\cite{shen2025fastvggttrainingfreeaccelerationvisual} & 0.0166 & 0.0042 & \textbf{0.0182} & \underline{0.0034} & 0.5554 & 0.5830 & 0.0181 & 0.0131& 0.0115& 0.0031& 0.7196& 0.8640\\
        SparseVGGT~\cite{wang2025sparsevggt} (SR: 50\%) & 0.0172 & 0.0039 & 0.0191 & 0.0042 & 0.5563 & 0.5846 & 0.0160 & 0.0099 & \textbf{0.0112} & \textbf{0.0027} & 0.7384 & 0.8787 \\
        SparseVGGT~\cite{wang2025sparsevggt} (SR: 75\%) & 0.0174 & 0.0040 & 0.0189 & 0.0042 & 0.5561 & 0.5842 & 0.0363 & 0.0258 & 0.0169 & 0.0041 & 0.6907 & 0.8353 \\
        Co-Me~\cite{chen2025comeconfidenceguidedtokenmerging} & \textbf{0.0147}& 0.0061& 0.0234&0.0060 &\textbf{0.5826} & \textbf{0.6271}& 0.0167 & 0.0091& 0.0115 & 0.0033& \textbf{0.7716} & \textbf{0.9104}\\
        LiteVGGT~\cite{shu2025litevggtboostingvanillavggt} & 0.0185 & 0.0059 & 0.0232 & \textbf{0.0033} & 0.5542 & 0.5815 & 0.0264 & 0.0154 & 0.0152 & 0.0030 & 0.6833 & 0.8009 \\
        \textbf{\methodname (Ours) ($\sigma=2$)} & \underline{0.0152}& \textbf{0.0036}& \underline{0.0188}& 0.0043& 0.5567& 0.5850&\underline{0.0127} & \underline{0.0075}&\textbf{0.0112} &\textbf{0.0027} & 0.7552 & 0.8946\\
        \textbf{\methodname (Ours) ($\sigma=3$)} & \underline{0.0152}& \textbf{0.0036}& 0.0189& 0.0043& \underline{0.5568}& \underline{0.5854}& \textbf{0.0126} & \textbf{0.0074} & 0.0113& 0.0028 & \underline{0.7582} & \underline{0.8973}\\
        \midrule
        $\pi^3$~\cite{wang2025pi3permutationequivariantvisualgeometry} \textit{(Base Model)} & \textcolor{gray}{0.0105}& \textcolor{gray}{0.0034}& \textcolor{gray}{0.0141}& \textcolor{gray}{0.0054}& \textcolor{gray}{0.5677}& \textcolor{gray}{0.6027}& \textcolor{gray}{0.0128}& \textcolor{gray}{0.0057} & \textcolor{gray}{0.0101}& \textcolor{gray}{0.0024}&\textcolor{gray}{0.7503} & \textcolor{gray}{0.8806}\\
        \midrule
        Sparse-$\pi^3$~\cite{wang2025sparsevggt} (SR: 50\%) & 0.0109 & 0.0036 & 0.0142 &0.0053  & 0.5664 & 0.6006 & 0.0150 & 0.0078 & 0.0112 & \textbf{0.0027} & 0.7361 & 0.8672 \\
        Sparse-$\pi^3$~\cite{wang2025sparsevggt} (SR: 75\%) & 0.0117 & 0.0038 & 0.0152 &0.0055  & 0.5659 & 0.5997 & 0.0189 & 0.0100 & 0.0149 & 0.0040 & 0.7350 & 0.8663 \\
        Speed3R~\cite{ren2026speed3r} & 0.0120 & 0.0040& \textbf{0.0137}& \textbf{0.0044}& 0.5661& 0.6006& 0.0208& 0.0137& 0.0149& 0.0040& 0.7256& 0.8617\\
        \textbf{\methodname (Ours) ($\sigma=2$)} & \underline{0.0105}& \underline{0.0034}& 0.0142& 0.0052& \textbf{0.5666}& \textbf{0.6009}& \textbf{0.0148}& \textbf{0.0074}& \textbf{0.0107}& \textbf{0.0027}& \textbf{0.7463}& \textbf{0.8781}\\
        \textbf{\methodname (Ours) ($\sigma=3$)} & \textbf{0.0104}& \textbf{0.0033}& \underline{0.0139}& \underline{0.0050}& \textbf{0.5666}& \underline{0.6008}& \underline{0.0149}& \underline{0.0073}& \textbf{0.0107}& \textbf{0.0027}& \underline{0.7461}& \underline{0.8778}\\
        \bottomrule[0.17em]
    \end{tabular}
    }
    \label{tab:mv_recon}
\end{table}



\subsection{Comparisons with existing methods}
\label{sec:compare_with_others}
\textbf{Camera pose estimation.} We evaluate our method on 7-Scenes, Neural RGB-D, and TUM-Dynamics. Following~\cite{cheng2025merg3r}, sequences in 7-Scenes are sampled every two frames, resulting in each scene containing 500 frames except for two scenes with 250 frames, while Neural RGB-D is sampled every five frames. As shown in \Cref{tab:relpose}, our method achieves an overall superior performance compared to existing approaches across all datasets, both on the long-sequence 7-Scenes and Neural RGB-D and the standard TUM-Dynamics benchmark, in several cases even improving upon the base model. These results highlight the effectiveness of our proposed token selection strategy, particularly as a \textit{training-free} approach, which can serve as a flexible plug-in that can be applied to general types of visual geometry transformers.



\textbf{3D point cloud reconstruction.} Unlike prior evaluations that often focus on sparse reconstruction from only 3-5 views per scene, we evaluate on dense multi-view settings to better assess the overall behavior on both performance and computational efficiency. Specifically, we follow the same protocol as in camera pose estimation and evaluate 3D point cloud reconstruction on 7-Scenes and Neural RGB-D with up to 500 frames per scene. As reported in \Cref{tab:mv_recon}, our method achieves superior overall performance compared to existing solutions, demonstrating its effectiveness and robustness in large-scale reconstruction scenarios.


\textbf{Video depth estimation.} To further evaluate performance on more tasks with long sequences, we conduct video depth estimation experiments on the full-length Bonn dataset~\cite{bonn}, where sequences range from 332 to 895 frames per scene. We adopt $\pi^3$ as the base model so that the baseline method can also operate within the 48GB memory. As shown in \Cref{tab:video_depth_bonn_full}, Sparse-$\pi^3$ encounters CUDA out-of-memory issues on these long sequences within 48GB memory, even with a high sparsification rate of 75\%. In contrast, our method scales reliably to sequences exceeding 800 frames, and even outperforms the base model. Moreover, compared to Speed3R which requires costly model retraining, our training-free approach still achieves superior performance on most metrics. 



\textbf{Inference efficiency.} We present the inference speed comparison across different methods in \Cref{tab:efficiency_comparison}. We can observe that our approach shows near-linear inference time scaling with respect to the number of input images, resulting in an improved efficiency for long sequences, as we select a constant budget of key/value tokens for the global attention layers. Although our method remains slightly slower than LiteVGGT, which achieves superior efficiency from expensive model retraining, our method consistently has smaller performance compromise. Combined with the comparisons on various tasks above, it demonstrates that our method achieves an overall better trade-off between efficiency and performance compared to existing approaches.



\begin{figure}[t]
\centering
\begin{minipage}[t]{0.55\textwidth}
    \rowcolors{2}{white}{uoftcoolgray!25}
    \centering
    \captionof{table}{Quantitative comparisons on video depth estimation. The best accelerated model is highlighted in \textbf{bold}. Our method outperforms the base model even in terms of quality.}
    \resizebox{\linewidth}{!}{
    \begin{tabular}{lccccc}
        \toprule[0.17em]
        \multicolumn{1}{l}{\multirow{2}{*}[-0.5ex]{\textbf{Method}}} &
        \multicolumn{5}{c}{\textbf{Bonn}} \\
        \cmidrule(r){2-6}
        \multicolumn{1}{c}{} &
        \textbf{Abs Rel ($\downarrow$)} & \textbf{Log RMSE ($\downarrow$)} & \textbf{RMSE ($\downarrow$)} & \textbf{Sq Rel ($\downarrow$)} & \textbf{$\delta<1.25$ ($\uparrow$)} \\
        \midrule[0.08em]
        $\pi^3$ \textit{(Base Model)} & \textcolor{gray}{0.0333} & \textcolor{gray}{0.0746} & \textcolor{gray}{0.1623} & \textcolor{gray}{0.0123} & \textcolor{gray}{0.9886} \\
        \midrule
        Sparse-$\pi^3$~\cite{wang2025sparsevggt} (SR: 50\%) & \textit{OOM} & \textit{OOM} & \textit{OOM} & \textit{OOM} & \textit{OOM} \\
        Sparse-$\pi^3$~\cite{wang2025sparsevggt} (SR: 75\%) & \textit{OOM} & \textit{OOM} & \textit{OOM} & \textit{OOM} & \textit{OOM} \\
        Speed3R~\cite{ren2026speed3r} & 0.0314 & 0.0680 & 0.1525 & 0.0103 & \textbf{0.9909} \\
        \textbf{\methodname (Ours) ($\sigma=3$)} & \textbf{0.0288} & \textbf{0.0668} & \textbf{0.1501} & \textbf{0.0100} & 0.9893 \\
        \bottomrule[0.17em]
    \end{tabular}
    }
    \label{tab:video_depth_bonn_full}
\end{minipage}
\hfill
\begin{minipage}[t]{0.43\textwidth}
    \rowcolors{2}{white}{uoftcoolgray!25}
    \centering
    \captionof{table}{Inference time comparison on an NVIDIA L40S GPU. As above, we use $K=25$ and $\sigma=3$. The fastest result is in \textbf{bold} and the second fastest is \underline{underlined}.}
    \resizebox{\linewidth}{!}{
        \begin{tabular}{lccccc}
        \toprule[0.17em]
        \textbf{Input Sequence Length}  & 
        \textbf{100} & 
        \textbf{200} & 
        \textbf{300} & \textbf{400} & \textbf{500}\\ 
        \midrule
        VGGT~\cite{wang2025vggtvisualgeometrygrounded} \textit{(Base Model)} & \textcolor{gray}{13.6s} & \textcolor{gray}{47.4s} & \textcolor{gray}{101.3s} & \textcolor{gray}{179.8s} & \textcolor{gray}{288.0s}\\
        \midrule
        FastVGGT~\cite{shen2025fastvggttrainingfreeaccelerationvisual} & 9.4s & 23.1s & 40.0s & 59.5s& 84.6s\\
        Sparse-$\pi^3$~\cite{wang2025sparsevggt} (SR: 50\%)& 6.8s & 18.3s & 35.0s & 56.2s&80.3s\\
        Sparse-$\pi^3$~\cite{wang2025sparsevggt} (SR: 75\%) & 5.7s & \underline{14.0s} & 25.6s & 39.5s &55.4s\\
        LiteVGGT~\cite{shu2025litevggtboostingvanillavggt} & \textbf{4.5s} & \textbf{10.1s} & \textbf{17.8s} & \textbf{26.0s} & \textbf{36.5s} \\
        Co-Me~\cite{chen2025comeconfidenceguidedtokenmerging} & \underline{5.5s} & 16.4s & 32.2s & 53.3s&84.2s\\
        \textbf{\methodname (Ours)} & 7.8s & 15.9s & \underline{23.8s} & \underline{31.7s} & \underline{41.2s}\\
        \bottomrule[0.17em]
        \end{tabular}
    }
    \label{tab:efficiency_comparison}
\end{minipage}

\end{figure}

\subsection{Ablation study and sensitivity analysis}
\label{sec:ablation}
Beyond the ablations already presented in Sections~\ref{sec:interframe} and~\ref{sec:intraframe} about the methodology design of both inter-frame and intra-frame strategies, we provide additional analysis below for the key parameters to offer practical guidance for applying our method.


\textbf{Budget for inter-frame selection.} In our default experimental setting, we use a fixed inter-frame budget of $K=25$ for scenes containing hundreds of frames. A natural concern is that using as few as 25 frames would be inevitably struggle to cover the full scene. To investigate this, we vary the budget $K$ allocated for inter-frame selection, and report the results in \Cref{tab:ab_inter_frame}. We observe that when the budget increases to 40-60 frames, which is rougly 10\% of the total frames, the performance gets further boosted compared to our default choice of $K=25$. What is more interesting is that the performance does not monotonically increase with larger budgets. On one hand, this is expected as when $K$ approaches the total number of input frames, the results should converge to the performance of the base model, which can be outperformed by many of our variants. On the other hand, it is still counterintuitive, as attending to more frames should, in principle, make the model visible to more information. 

There are many factors that can possibly cause this counterintuitive phenomena to happen, below are two of our proposed potential reasons: \textit{\textbf{(1) More selected frames indicate more retained tokens for global attention}}, and this may potentially aggravate attention dilution if this issue exists. This is consistent with our observation and solution for intra-frame token selection. As illustrated in~\Cref{fig:attention_pattern}, early layers show diluted, near-uniform attention patterns, and downsampling key/value tokens that suffer from dilution can help improve the performance (\Cref{tab:different_layer_intra_frame_avggt} when downsampling the heavily diluted 0-8 layers). Similarly, if the tokens in the original VGGT suffer from a certain level of attention dilution, then the inter-frame downsampling is also alleviating the dilution issue, which accounts for the non-monotonic behavior of the inter-frame budget $K$. \textit{\textbf{(2) The base models VGGT and $\pi^3$ are trained with the maximum sequence length of $l=24$}}, and demonstrate the generalization capability to longer sequences containing hundreds of frames. However, it is possible that the generalization capability can gradually decrease when the number of input frames grows, and the bottleneck for the generalization capability lies in the global attention layer that we are optimizing. Therefore, when the number of frames grows, on one hand, the available multi-view information increases for cross-view information aggregation; while on the other hand, the generalization capability may gradually diminish as the number of frames increases. These two factors may ultimately reach a "sweet spot" at $K$ being around 40-60, as demonstrated from the results in~\Cref{tab:ab_inter_frame}, which causes the non-monotonic behavior of $K$.

\textbf{Layer thresholds for intra-frame selection.} In~\Cref{sec:intraframe}, we set the thresholds $l_{\rm local}$ and $l_{\rm sample}$ to determine the layer-adaptive strategies with different levels of token pruning across all the global attention layers. We vary these thresholds in \Cref{tab:ab_intra_frame} and find that the performance remains stable across a broad range of configurations, which is consistent with the observed global attention patterns for each layer. This indicates that our method is robust to these hyperparameter choices, which further enhances the soundness and reliability of our method.


\begin{figure}[t]
\centering
\begin{minipage}[t]{0.55\textwidth}
        \rowcolors{2}{white}{uoftcoolgray!25}
    \centering
    \captionof{table}{Analysis on the impact of the inter-frame selection budget $K$ with $\sigma=3$ for the pose estimation task on 7-Scenes. Increasing $K$ initially improves performance, indicating that a moderate expansion of the selected frame set enhances scene coverage. However, the trend is not monotonic: beyond a certain range, further increasing $K$ occasionally leads to an eventual performance degradation, with the performance approaching the results of the full model. Inference time is measured on an NVIDIA L40S GPU with 500 frames.
}
    \resizebox{\linewidth}{!}{
        \begin{tabular}{ccccc}
        \toprule[0.17em]
        \textbf{$K$}  & 
        \textbf{ATE ($\downarrow$)} & 
        \textbf{RPE-rot ($\downarrow$)} & 
        \textbf{RPE-trans ($\downarrow$)} & \textbf{Inference Time ($\downarrow$)} \\ 
        \midrule
        VGGT \textit{(Base Model)} & \textcolor{gray}{0.0698} & \textcolor{gray}{0.4953} & \textcolor{gray}{0.0178} & \textcolor{gray}{288.0s}\\
        \midrule
        10 & 0.0722 & 0.7614 & 0.0198 & 32.3s\\
        25 & 0.0677 & 0.4495 & 0.0166 & 41.2s\\
        40 & 0.0674 & 0.4204 & 0.0155 & 51.9s\\
        60 & 0.0677 & 0.4203 & 0.0153 & 63.7s\\
        80 & 0.0684 & 0.4211 & 0.0152 & 77.8s\\
        100 & 0.0685 & 0.4229 & 0.0153& 89.3s\\
        \midrule
        $\pi^3$ \textit{(Base Model)} & \textcolor{gray}{0.0573} & \textcolor{gray}{0.3389} & \textcolor{gray}{0.0105} & \textcolor{gray}{110.1s}\\
        \midrule
        10 & 0.0575 & 0.3652 & 0.0126 & 21.9s\\
        25 & 0.0570 & 0.3428 & 0.0112 & 26.1s\\
        40 & 0.0563 & 0.3384 & 0.0108 & 30.5s\\
        60 & 0.0561 & 0.3379 & 0.0107 & 36.1s\\
        80 & 0.0567 & 0.3348 & 0.0105 & 42.2s\\
        100 & 0.0566 & 0.3359 & 0.0106& 47.8s\\
        \bottomrule[0.17em]
        \end{tabular}
    }
    \label{tab:ab_inter_frame}
\end{minipage}
\hfill
\begin{minipage}[t]{0.43\textwidth}
    \rowcolors{2}{white}{uoftcoolgray!25}
    \centering
    \captionof{table}{Sensitivity analysis on the layer partition thresholds $l_{\rm local}$ and $l_{\rm sample}$ for intra-frame selection with $K=25$, with the \textit{\textcolor{uoftsecondaryblue}{blue}} row indicating our current parameter choice. Performance remains stable across a wide range of threshold choices, demonstrating that our method is robust to these hyperparameter choices, as long as they are consistent with the observed layer-wise attention patterns.}
    \resizebox{\linewidth}{!}{
        \begin{tabular}{ccccc}
        \toprule[0.17em]
        $l_{\rm local}$  & $l_{\rm sample}$  & 
        \textbf{ATE ($\downarrow$)} & 
        \textbf{RPE-rot ($\downarrow$)} & 
        \textbf{RPE-trans ($\downarrow$)} \\ 
        \midrule
        \multicolumn{2}{l}{$\pi^3$ \textit{(Base Model)}} & \textcolor{gray}{0.0573} & \textcolor{gray}{0.3389} & \textcolor{gray}{0.0105}\\
        \midrule
        1 & 8 & 0.0567 & 0.3425 & 0.0112\\
        1 & 9 & 0.0568 & 0.3425 & 0.0112\\
        1 & 10 & 0.0569 & 0.3406 & 0.0111\\
        2 & 8 & 0.0569 & 0.3432 & 0.0112\\
        \color{uoftsecondaryblue}2 & \color{uoftsecondaryblue}9 & \color{uoftsecondaryblue}0.0570 & \color{uoftsecondaryblue}0.3428 & \color{uoftsecondaryblue}0.0112\\
        2 & 10 & 0.0570 & 0.3409 & 0.0112\\
        3 & 8 & 0.0570 & 0.3415 & 0.0112\\
        3 & 9 & 0.0571 & 0.3418 & 0.0112\\
        3 & 10 & 0.0571 & 0.3396 & 0.0112\\
        4 & 8 & 0.0577 & 0.3442 & 0.0113\\
        4 & 9 & 0.0578 & 0.3446 & 0.0113\\
        4 & 10 & 0.0578 & 0.3425 & 0.0113\\
        \bottomrule[0.17em]
        \end{tabular}
    }
    \label{tab:ab_intra_frame}
\end{minipage}

\end{figure}



%% file: secs/5_discussion.tex
\section{Discussions}
\label{sec:discussions}

Although our approach is introduced as a training-free acceleration method, the analysis and findings in our paper point to a more fundamental observation: the improved performance after token selection indicates that current visual geometry transformers are not yet perfectly trained with the optimal architecture. Our solution thus provides guidelines for future research on how to improve the network design and training strategies of visual geometry transformers. For example, our study on inter-frame selection points toward the potential of routing-based mechanisms for attention layers inside these models, while our intra-frame analysis suggests that global attention layers at early stages which suffer from attention dilution may be skipped even in the training process.

\textbf{Limitations.} Our inter-frame selection strategy relies on features extracted by a place recognition model, which may become less reliable in challenging scenarios such as object-centric scenes or environments with highly symmetric and ambiguous structures. However, these settings are inherently difficult for visual geometry transformers in general, and remain challenging even for conventional optimization-based structure-from-motion and multi-view stereo pipelines. In addition, the capability of our approach may get bounded by the capacity of the underlying base model. While our method substantially improves efficiency through hierarchical token selection, it does not alter the overall inference paradigm of visual geometry transformers. Consequently, it still faces difficulties on extremely large-scale environments, such as kilometer-scale scenes considered in VGGT-Long~\cite{deng2025vggtlongchunkitloop}, where additional techniques such as chunk-based processing or recurrent inference are required. Nevertheless, our method is orthogonal and complementary to these approaches, and can still potentially be integrated with them to further improve their efficiency.




%% file: secs/6_conclusion.tex
\section{Conclusions}
\label{sec:conclusions}

We present \methodname, a study that formulates the efficiency improvement of visual geometry transformers as a token selection problem. To enable effective selection, we introduce a two-stage hierarchical framework consisting of inter-frame and intra-frame selection. Through comprehensive analysis, we show that diversity-based strategies are well-suited for inter-frame selection, while layer-adaptive strategies with different levels of token pruning are needed for intra-frame selection. Extensive experiments demonstrate that our method achieves a superior trade-off between inference efficiency and reconstruction quality compared to existing approaches, even occasionally
outperforming the base models. We believe that our training-free solution can serve as a general and easy-to-use algorithm for accelerating various visual geometry transformers. 



%% file: checklist.tex
\section*{NeurIPS Paper Checklist}

\begin{enumerate}

\item {\bf Claims}
    \item[] Question: Do the main claims made in the abstract and introduction accurately reflect the paper's contributions and scope?
    \item[] Answer: \answerYes{} 
    \item[] Justification: The claims made in the abstract and introduction accurately reflect the paper's
contributions and scope.

\item {\bf Limitations}
    \item[] Question: Does the paper discuss the limitations of the work performed by the authors?
    \item[] Answer: \answerYes{} 
    \item[] Justification:  The limitations of the work have been discussed in~\Cref{sec:discussions} in the main paper.

\item {\bf Theory assumptions and proofs}
    \item[] Question: For each theoretical result, does the paper provide the full set of assumptions and a complete (and correct) proof?
    \item[] Answer: \answerNA{} 
    \item[] Justification: The paper does not include theoretical results.

    \item {\bf Experimental result reproducibility}
    \item[] Question: Does the paper fully disclose all the information needed to reproduce the main experimental results of the paper to the extent that it affects the main claims and/or conclusions of the paper (regardless of whether the code and data are provided or not)?
    \item[] Answer: \answerYes{} 
    \item[] Justification: We have discussed all the details to reproduce all experimental results in our
paper.

\item {\bf Open access to data and code}
    \item[] Question: Does the paper provide open access to the data and code, with sufficient instructions to faithfully reproduce the main experimental results, as described in supplemental material?
    \item[] Answer: \answerYes{} 
    \item[] Justification: We have published our code at \url{https://github.com/zsh2000/gotohunt}.
    

\item {\bf Experimental setting/details}
    \item[] Question: Does the paper specify all the training and test details (e.g., data splits, hyperparameters, how they were chosen, type of optimizer) necessary to understand the results?
    \item[] Answer: \answerYes{} 
    \item[] Justification: The inference details including data processing and hyperparameter choices have been specified in the paper, which are enough to understand the results.

\item {\bf Experiment statistical significance}
    \item[] Question: Does the paper report error bars suitably and correctly defined or other appropriate information about the statistical significance of the experiments?
    \item[] Answer: \answerYes{} 
    \item[] Justification: We have measured the variance of different seeds in~\Cref{sec:variance_across_seeds} in the Appendix.

\item {\bf Experiments compute resources}
    \item[] Question: For each experiment, does the paper provide sufficient information on the computer resources (type of compute workers, memory, time of execution) needed to reproduce the experiments?
    \item[] Answer: \answerYes{} 
    \item[] Justification:  The compute resources required for reproduce the is specified in~\Cref{sec:experiments} in the paper.
    
\item {\bf Code of ethics}
    \item[] Question: Does the research conducted in the paper conform, in every respect, with the NeurIPS Code of Ethics \url{https://neurips.cc/public/EthicsGuidelines}?
    \item[] Answer: \answerYes{} 
    \item[] Justification:  The conducted research conforms with the NeurIPS Code of Ethics in every
respect.

\item {\bf Broader impacts}
    \item[] Question: Does the paper discuss both potential positive societal impacts and negative societal impacts of the work performed?
    \item[] Answer: \answerYes{} 
    \item[] Justification: Both positive and negative societal impacts have been discussed in~\Cref{sec:Societal} in
the Appendix.
    
\item {\bf Safeguards}
    \item[] Question: Does the paper describe safeguards that have been put in place for responsible release of data or models that have a high risk for misuse (e.g., pre-trained language models, image generators, or scraped datasets)?
    \item[] Answer: \answerNA{} 
    \item[] Justification: Our paper does not have such risks.

\item {\bf Licenses for existing assets}
    \item[] Question: Are the creators or original owners of assets (e.g., code, data, models), used in the paper, properly credited and are the license and terms of use explicitly mentioned and properly respected?
    \item[] Answer: \answerYes{} 
    \item[] Justification: We have properly credited and listed the license in~\Cref{sec:licenses} in the Appendix. Please note that licenses for some datasets are extremely lengthy as each scene in these datasets have different licenses. Therefore, we include the links for reference instead of directly listing all the licenses. 

\item {\bf New assets}
    \item[] Question: Are new assets introduced in the paper well documented and is the documentation provided alongside the assets?
    \item[] Answer: \answerNA{} 
    \item[] Justification: At the submission time, we have not released the new assets used in the paper. 

\item {\bf Crowdsourcing and research with human subjects}
    \item[] Question: For crowdsourcing experiments and research with human subjects, does the paper include the full text of instructions given to participants and screenshots, if applicable, as well as details about compensation (if any)? 
    \item[] Answer: \answerNA{} 
    \item[] Justification: The paper does not involve crowdsourcing nor research with human subjects.

\item {\bf Institutional review board (IRB) approvals or equivalent for research with human subjects}
    \item[] Question: Does the paper describe potential risks incurred by study participants, whether such risks were disclosed to the subjects, and whether Institutional Review Board (IRB) approvals (or an equivalent approval/review based on the requirements of your country or institution) were obtained?
    \item[] Answer: \answerNA{} 
    \item[] Justification:  the paper does not involve crowdsourcing nor research with
human subjects.

\item {\bf Declaration of LLM usage}
    \item[] Question: Does the paper describe the usage of LLMs if it is an important, original, or non-standard component of the core methods in this research? Note that if the LLM is used only for writing, editing, or formatting purposes and does \emph{not} impact the core methodology, scientific rigor, or originality of the research, declaration is not required.
    \item[] Answer: \answerNA{} 
    \item[] Justification:  The core method development in this research does not involve LLMs as any
important, original, or non-standard components.

\end{enumerate}

%% file: secs/X_supp.tex
{\Large\centering\bf 	
Good Token Hunting: A Hitchhiker's Guide to Token Selection for Visual Geometry Transformers\\}

\vspace{4mm}
{\large\centering\rmfamily Technical Appendices and Supplementary Material\\}

\vspace{4mm}


In the appendix, we provide additional clarfication, analyses and experiments to further validate the effectiveness and provide guidelines of our method along with more discussions. First, in~\Cref{sec:inter_frame_alg} and~\Cref{sec:variance_across_seeds}, we illustrate the details of the proposed diversity-based inter-frame selection algorithm, and the robustness of random seed selection in this process.~\Cref{sec:additional_intra_frame} displays the exploration of several intra-frame strategies that we attempted during our research, but ultimately chose not to be applied on our final method, in order to provide more guidelines for future research. Then, in~\Cref{sec:additional_sensitivity}, we provide additional experiments on the sensitivity analysis of the hyperparameters used in our method, demonstrating the robustness of our method to parameter choices. In addition, we illustrate in~\Cref{sec:other_vggt} that our method can be easily applied to other visual geometry transformers with different architecture.
Afterwards, in~\Cref{sec:evaluation_metrics}, we elaborate the evaluation metrics used in the tasks.~\Cref{sec:licenses} include the license and terms of use for the models and data used in the paper. Finally,~\Cref{sec:additional_discussion} and~\Cref{sec:Societal} provide additional discussions and societal impact of our work.


\section{Algorithm details for inter-frame selection}
\label{sec:inter_frame_alg}

In~\Cref{sec:interframe}, we propose the diversity-based inter-frame selection solution. We demonstrate the algorithm details in~\Cref{alg:diverse-frames}. Given image-level features extracted by a place recognition model~\cite{Berton_2025_MegaLoc}, we first construct a pairwise similarity matrix using cosine similarity, serving as approximation for co-visibility between frames, which is then converted into a distance metric for the following sampling process. The selection problem is formulated as a "$K$-center" objective, aiming to choose a subset of frames that maximizes coverage of the scene. To efficiently approximate this objective, we adopt the greedy farthest point sampling (FPS) algorithm. Starting from a randomly initialized frame, FPS iteratively selects the frame that is farthest from the current selected set, thereby encouraging diversity among selected frames. The resulting subset is shared across all query tokens and serves as the inter-frame support "anchor" set for global attention. In practice, this procedure is computationally efficient, requiring only a single $N \times N$ similarity computation and $K$ iterative updates. 

Additionally, we would like to point out that for VGGT~\cite{wang2025vggtvisualgeometrygrounded}, the selected reference set needs to include the first frame, which serves as the reference view of the scene from the model design. In our implementation, if the above selection process is done but the first frame is still absent in the selected set, we replace the final selected frame with the first frame to preserve the budget $K$. No corresponding rules for the reference need to be applied for the permutation-equivariant architectures like $\pi^3$~\cite{wang2025pi3permutationequivariantvisualgeometry}.

\renewcommand{\algorithmicrequire}{\textbf{Input:}}
\renewcommand{\algorithmicensure}{\textbf{Output:}}

\begin{algorithm}[ht]
\caption{Diversity-based Inter-frame Selection}
\label{alg:diverse-frames}
\begin{algorithmic}[1]
\Require Feature map $F \in \mathbb{R}^{N \times d}$ (row $f_i$ is the $d$-dim feature of image $i$);
         number of neighbors $K$;
         random seed $\sigma$
\Ensure  Selected index set $S$ with $|S| = K$ (shared by all query frames)
\Statex
\State \textbf{// Step 1: build the covisibility matrix from features (cosine similarity).}

\State $\tilde{f}_i \gets f_i / \|f_i\|_2, \quad \forall\, i \in \{0,\dots,N-1\}$
       \Comment{$\mathcal{L}_2$-normalized each row}
\State $\tilde{F} \gets [\,\tilde{f}_0;\; \tilde{f}_1;\; \dots;\; \tilde{f}_{N-1}\,] \in \mathbb{R}^{N \times d}$
       \Comment{stack normalized rows}
\State $C \gets \tilde{F}\tilde{F}^{\top}, \quad C_{i,j} \in [-1, 1]$ \Comment{covisibility matrix with cosine similarity between features}
       
\Statex
\State \textbf{// Step 2: Convert covisibility to a distance.}
\State $D_{i,j} \gets \max(C) - C_{i,j}, \quad \forall\, i,j$
       \Comment{larger covisibility $\Rightarrow$ closer distance}
\Statex
\State \textbf{// Step 3: FPS starts from a random first pick.}
\State $b_1 \sim \mathrm{Uniform}\{0,\dots,N-1\}$ \Comment{random selection with seed $\sigma$}
\State $S \gets \{b_1\}$ \Comment{add frame to the selected set}
\State $d_{\min} \gets D_{b_1,:}$ \Comment{$d_{\min}\in\mathbb{R}^N$ stores the distance of each frame to the selected set}
\State $d_{\min}[b_1] \gets -\infty$  \Comment{prevent re-selecting the already selected frame}
\Statex
\State \textbf{// Step 4: FPS for the remaining $K-1$ picks.}
\For{$k \gets 2$ \textbf{to} $K$}
    \State $b \gets \arg\max_{j} d_{\min}[j]$  \Comment{select the farthest frame from the currently selected set}
    \State $S \gets S \cup \{b\}$ \Comment{add frame to the selected set}
    \State $d_{\min}[b] \gets -\infty$ \Comment{prevent re-selecting the already selected frame}
    \State $d_{\min} \gets \min\!\big(d_{\min},\; D_{b,:}\big)$   \Comment{update the distance change due to the newly added frame}
\EndFor
\Statex
\State \Return $S$
\end{algorithmic}
\end{algorithm}

\section{Variance analysis on random seeds for inter-frame selection}
\label{sec:variance_across_seeds}

In~\Cref{sec:inter_frame_alg}, we mention that inter-frame selection starts from a randomly initialized frame and performs the FPS algorithm afterwards. To demonstrate the robustness of model performance regarding random seed selection, we use 5 random seeds and report the mean and standard deviation for the camera pose estimation task in~\Cref{tab:seed_variance_unverified}. We can observe that the performance of our method is generally robust to the randomness from the first frame pick. This randomness stems from the nature of our diversity-based frame selection algorithm. As long as the number of selected frames 
is not too small to cover most parts of the 3D scene, the selected frames from our~\Cref{alg:diverse-frames}, inspired by farthest point sampling, can excel in finding a set of frames that cover most of the 3D scene to achieve satisfactory results.

\begin{table}[ht]
\centering
\rowcolors{2}{white}{uoftcoolgray!25}
\caption{Quantitative results camera pose estimation on 5 random seeds with mean and standard deviation. $K=25$ is selected for the experimental settings.}
\vspace{5pt}
\resizebox{0.65\linewidth}{!}{%
\begin{tabular}{llccc}
\toprule[0.17em]
\textbf{Backbone} & \textbf{$\sigma$} & \textbf{ATE ($\downarrow$)} & \textbf{RPE-rot ($\downarrow$)} & \textbf{RPE-trans ($\downarrow$)} \\
\midrule
\rowcolor{white}\multicolumn{5}{c}{\textbf{7-Scenes}} \\ \midrule
VGGT & 2 & 0.0669 $\pm$ 0.0009 & 0.4372 $\pm$ 0.0135 & 0.0164 $\pm$ 0.0003 \\
VGGT & 3 & 0.0672 $\pm$ 0.0009 & 0.4386 $\pm$ 0.0153 & 0.0165 $\pm$ 0.0004 \\
$\pi^3$ & 2 & 0.0576 $\pm$ 0.0006 & 0.3447 $\pm$ 0.0019 & 0.0112 $\pm$ 0.0001 \\
$\pi^3$ & 3 & 0.0576 $\pm$ 0.0006 & 0.3441 $\pm$ 0.0018 & 0.0112 $\pm$ 0.0001 \\
\midrule
\rowcolor{white}\multicolumn{5}{c}{\textbf{Neural RGB-D}} \\ \midrule
VGGT & 2 & 0.0273 $\pm$ 0.0010 & 0.1946 $\pm$ 0.0464 & 0.0161 $\pm$ 0.0003 \\
VGGT & 3 & 0.0272 $\pm$ 0.0009 & 0.2109 $\pm$ 0.0231 & 0.0175 $\pm$ 0.0005 \\
$\pi^3$ & 2 & 0.0292 $\pm$ 0.0011 & 0.1184 $\pm$ 0.0016 & 0.0122 $\pm$ 0.0001 \\
$\pi^3$ & 3 & 0.0296 $\pm$ 0.0009 & 0.1199 $\pm$ 0.0018 & 0.0122 $\pm$ 0.0001 \\
\bottomrule[0.17em]
\end{tabular}}
\label{tab:seed_variance_unverified}
\end{table}

\section{Additional analysis on intra-frame strategy}
\label{sec:additional_intra_frame}

\subsection{Token-level diversity-based selection}
\label{sec:tld}

In~\Cref{sec:intraframe}, we adopt an intra-frame strategy to downsample the token map along the two spatial dimensions, which is independent to the token content. One intuitive consideration is that, what if we adopt a similar diversity-based strategy as inter-frame selection, in the current intra-frame stage, leading to a diversity-based intra-frame selection strategy. However, the total number of tokens across frames can be prohibitively large for directly applying such a method even after inter-frame selection. To address this, we design an approximate solution. Specifically, for each selected frame, we first over-select twice the target budget using FPS. Then, we compute a \textit{redundancy score} for each token, which is its maximum cosine similarity to the mean feature of every other frame. Finally, we retain the least redundant tokens from the selected tokens. This procedure directly guarantees diversity within each frame and, to some extent, promotes token diversity across frames. We refer to this Token-level Diversity-based strategy as \textit{TLD}. 

We observe that \textit{TLD} performs well in certain scenarios, such as video depth estimation in~\Cref{tab:supp_intra_frame_video_depth}, where it consistently outperforms the \textit{standard} strategy in nearly all configurations. However, in other tasks, including pose estimation in~\Cref{tab:supp_intra_frame_pose} and 3D reconstruction in~\Cref{tab:supp_intra_frame_recon}, its performance is comparable to or sometimes worse than the \textit{standard} approach.

In addition, unlike diversity-based inter-frame sampling, which is performed only once in the image space, \textit{TLD} requires FPS within each selected frame even before advancing to the second selection stage, resulting in non-negligible computational overhead. It typically takes roughly 5 seconds for a scene with 500 frames. Therefore, this strategy is not applied in our official solution, but we report it here to potentially inspire future research.

\begin{table}[ht]
\centering
\rowcolors{2}{white}{uoftcoolgray!25}
\caption{Quantitative comparison on the token-level diversity-based \textit{(TLD)} selection strategy for intra-frame downsampling with video depth prediction on the full length Bonn~\cite{bonn} dataset.}
\vspace{5pt}
\resizebox{0.85\linewidth}{!}{
\begin{tabular}{cccccccc}
\toprule[0.17em]
\textbf{$K$} & \textbf{$\sigma$} & \textbf{Strategy} & \textbf{Layers} & \textbf{Abs Rel ($\downarrow$)} & \textbf{Log RMSE ($\downarrow$)} & \textbf{RMSE ($\downarrow$)} & \textbf{$\delta < 1.25$ ($\uparrow$)} \\
\midrule
10 & 2 & \textit{TLD} & 0-10 & 0.0305 & 0.0686 & 0.1512 & 0.9889 \\
10 & 2 & \textit{Standard} & 0-10 & 0.0352 & 0.0748 & 0.1593 & 0.9882 \\
10 & 3 & \textit{TLD} & 0-10 & 0.0292 & 0.0666 & 0.1480 & 0.9892 \\
10 & 3 & \textit{Standard}   & 0-10 & 0.0346 & 0.0717 & 0.1553 & 0.9892 \\
10 & 3 & \textit{TLD} & 11-17 & 0.0416 & 0.0779 & 0.1653 & 0.9841 \\
10 & 3 & \textit{Standard}   & 11-17 & 0.0436 & 0.0815 & 0.1675 & 0.9824 \\

\midrule

25 & 2 & \textit{TLD} & 0-10 & 0.0290 & 0.0667 & 0.1484 & 0.9892 \\
25 & 2 & \textit{Standard}   & 0-10 & 0.0340 & 0.0729 & 0.1565 & 0.9887 \\
25 & 3 & \textit{TLD} & 0-10 & 0.0290 & 0.0651 & 0.1457 & 0.9896 \\
25 & 3 & \textit{Standard}   & 0-10 & 0.0303 & 0.0664 & 0.1481 & 0.9897 \\

\midrule

100 & 2 & \textit{TLD} & 0-10 & 0.0283 & 0.0666 & 0.1495 & 0.9895 \\
100 & 2 & \textit{Standard}   & 0-10 & 0.0336 & 0.0729 & 0.1568 & 0.9889 \\
100 & 3 & \textit{TLD} & 0-10 & 0.0275 & 0.0648 & 0.1605 & 0.9901 \\
100 & 3 & \textit{Standard}   & 0-10 & 0.0291 & 0.0653 & 0.1702 & 0.9901 \\

\midrule
\multicolumn{4}{l}{$\pi^3$ \textit{(Base Model)}} & \textcolor{gray}{0.0346} & \textcolor{gray}{0.0763}& \textcolor{gray}{0.1642}& \textcolor{gray}{0.9880}\\
\bottomrule[0.17em]
\end{tabular}
}
\label{tab:supp_intra_frame_video_depth}
\end{table}

\begin{table}[ht]
\centering
\rowcolors{2}{white}{uoftcoolgray!25}
\caption{Quantitative comparison on the token-level diversity-based \textit{(TLD)} selection strategy for intra-frame downsampling with camera pose estimation on 7-Scenes~\cite{7scenes} dataset.}
\vspace{5pt}
\resizebox{0.75\linewidth}{!}{
\begin{tabular}{ccccccc}
\toprule[0.17em]
\textbf{$K$} & \textbf{$\sigma$} & \textbf{Strategy} & \textbf{Layers} & \textbf{ATE ($\downarrow$)} & \textbf{RPE-rot ($\downarrow$)} & \textbf{RPE-trans ($\downarrow$)} \\
\midrule
10 & 2 & \textit{TLD} & 0-10 & 0.0578 & 0.3602 & 0.0123 \\
10& 2 &  \textit{Standard}& 0-10 & 0.0589 & 0.3675 & 0.0125 \\
10 & 3 & \textit{TLD} & 0-10 & 0.0581 & 0.3584 & 0.0122 \\
10 & 3 & \textit{Standard} & 0-10 & 0.0587 & 0.3747 & 0.0130 \\
 \midrule
 25 & 2 & \textit{TLD} & 0-10 & 0.0576 & 0.3423 & 0.0111 \\
25 & 2 & \textit{Standard} & 0-10 & 0.0580 & 0.3430 & 0.0112 \\
25 & 3 & \textit{TLD} & 0-10 & 0.0570 & 0.3415 & 0.0111 \\
25 & 3 & \textit{Standard} & 0-10 & 0.0568 & 0.3415 & 0.0112 \\
\midrule
100 & 2 & \textit{TLD} & 0-10 & 0.0573 & 0.3363 & 0.0105 \\
100 & 2 & \textit{Standard} & 0-10 & 0.0577 & 0.3369 & 0.0106 \\
100 & 3 & \textit{TLD} & 0-10 & 0.0571 & 0.3384 & 0.0106 \\
100 & 3 & \textit{Standard} & 0-10 & 0.0566 & 0.3384 & 0.0106 \\
\midrule
\multicolumn{4}{l}{$\pi^3$ \textit{(Base Model)}} & \textcolor{gray}{0.0573} & \textcolor{gray}{0.3389}& \textcolor{gray}{0.0105}\\
\bottomrule[0.17em]
\end{tabular}
}
\label{tab:supp_intra_frame_pose}
\end{table}

\begin{table}[ht]
\centering
\rowcolors{2}{white}{uoftcoolgray!25}
\caption{Quantitative comparison on the token-level diversity-based \textit{(TLD)} selection strategy for intra-frame downsampling with point map estimation on Neural RGB-D~\cite{Azinovic_2022_CVPR} dataset.}
\vspace{5pt}
\resizebox{0.65\linewidth}{!}{
\begin{tabular}{ccccccc}
\toprule[0.17em]
\textbf{$K$} & \textbf{$\sigma$} & \textbf{Strategy} & \textbf{Layers} & \textbf{Acc ($\downarrow$)} & \textbf{Comp ($\downarrow$)} & \textbf{NC ($\uparrow$)} \\
\midrule
10 & 2 &  \textit{TLD} & 0-8 & 0.0158 & 0.0119 & 0.7486 \\
10 & 2 & \textit{Standard} & 0-8 & 0.0153 & 0.0114 & 0.7477 \\
10 & 3 & \textit{TLD} & 0-8 & 0.0161 & 0.0119 & 0.7485 \\
10 & 3 & \textit{Standard} & 0-8 & 0.0145 & 0.0112 & 0.7470 \\
\midrule
25 & 2 &  \textit{TLD} & 0-8 & 0.0126 & 0.0112 & 0.7588 \\
25 & 2 & \textit{Standard} & 0-8 & 0.0126 & 0.0110 & 0.7561 \\
25 & 3 &  \textit{TLD} & 0-8 & 0.0127 & 0.0111 & 0.7571 \\
25 & 3 & \textit{Standard} & 0-8 & 0.0126 & 0.0111 & 0.7564 \\
\midrule
100 & 2 &  \textit{TLD} & 0-8 & 0.0128 & 0.0113 & 0.7609 \\
100 & 2 & \textit{Standard} & 0-8 & 0.0130 & 0.0111 & 0.7596 \\
100 & 3 &  \textit{TLD} & 0-8 & 0.0130 & 0.0113 & 0.7602 \\
100 & 3 & \textit{Standard} & 0-8 & 0.0129 & 0.0112 & 0.7626 \\
\midrule
\multicolumn{4}{l}{VGGT \textit{(Base Model)}} & \textcolor{gray}{0.0160} & \textcolor{gray}{0.0112}& \textcolor{gray}{0.7508} \\
\bottomrule[0.17em]
\end{tabular}
}
\label{tab:supp_intra_frame_recon}
\end{table}

\subsection{Replacing global attention with mean pooling}

Replacing global attention with local attention adopted in the main paper is already a fairly radical strategy. An even more aggressive alternative is to skip the attention mechanism entirely and replace it with mean pooling over all value tokens. The justification of this simplification comes from the attention formulation, where the attention output for each query $q_i$ is $o_i=\sum_{j}{\rm softmax}{\left(\frac{q_i\cdot k_j}{\sqrt{d}}\right)}v_j$. If the attention activations $q_i\cdot k_j$ are nearly identical across all $j$, then the softmax weights approach $\frac{1}{N}$, where $N$ is the number of total K/V tokens. In this regime, the attention calculation collapse to a simple averaging of the all value tokens.

We implement this variant with the results reported in~\Cref{tab:vggt_pool_strategy}. The performance degradation is noticeable across many configurations, suggesting that this approximation is too crude and is not robust enough for accelerating visual geometry transformers.


\begin{table}[ht]
\rowcolors{2}{white}{uoftcoolgray!25}
\centering
\captionof{table}{Quantitative comparison on the more extreme approximation to replace global attention layers with mean pooling. The experiments are conducted with $K=25$ and $\sigma=2$ on the camera pose estimation on Neural RGB-D~\cite{Azinovic_2022_CVPR}. Essentially, we replace the local attention \textit{(Local)} in $l\leqslant l_{\rm local}$ with mean pooling \textit{(Pool)} layers. The inferior performance with the \textit{Pool} strategy demonstrates that it is not robust enough to be instantiated in our method.}
\vspace{5pt}
\resizebox{0.7\linewidth}{!}{
\begin{tabular}{cccccc}
\toprule[0.17em]
$l_{\rm local}$ & $l_{\rm global}$ & \textbf{Strategy} &
\textbf{ATE ($\downarrow$)} &
\textbf{RPE-rot ($\downarrow$)} &
\textbf{RPE-trans ($\downarrow$)} \\
\midrule
1 & 7 & \textit{Local} & 0.0268 & 0.2118 & 0.0170 \\
1 & 7 & \textit{Pool}     & 0.0288 & 0.3353 & 0.0198 \\
\midrule
1 & 8 & \textit{Local} & 0.0264 & 0.1946 & 0.0165 \\
1 & 8 & \textit{Pool}     & 0.0286 & 0.3235 & 0.0195 \\
\midrule
1 & 9 & \textit{Local} & 0.0266 & 0.1907 & 0.0164 \\
1 & 9 & \textit{Pool}     & 0.0286 & 0.3163 & 0.0193 \\
\midrule
1 & 10 & \textit{Local} & 0.0266 & 0.1659 & 0.0158 \\
1 & 10 & \textit{Pool}     & 0.0280 & 0.2809 & 0.0184 \\
\midrule
2 & 7 & \textit{Local} & 0.0269 & 0.2000 & 0.0169 \\
2 & 7 & \textit{Pool}     & 0.0324 & 0.4391 & 0.0242 \\
\midrule
2 & 8 & \textit{Local} & 0.0266 & 0.1847 & 0.0163 \\
2 & 8 & \textit{Pool}     & 0.0316 & 0.4261 & 0.0236 \\
\midrule
2 & 9 & \textit{Local} & 0.0267 & 0.1794 & 0.0162 \\
2 & 9 & \textit{Pool}     & 0.0318 & 0.4248 & 0.0235 \\
\midrule
2 & 10 & \textit{Local} & 0.0267 & 0.1615 & 0.0158 \\
2 & 10 & \textit{Pool}     & 0.0310 & 0.3936 & 0.0223 \\
\midrule
3 & 7 & \textit{Local} & 0.0268 & 0.1652 & 0.0160 \\
3 & 7 & \textit{Pool}     & 0.0335 & 0.4444 & 0.0247 \\
\midrule
3 & 8 & \textit{Local} & 0.0266 & 0.1548 & 0.0158 \\
3 & 8 & \textit{Pool}     & 0.0328 & 0.4336 & 0.0240 \\
\midrule
3 & 9 & \textit{Local} & 0.0267 & 0.1538 & 0.0156 \\
3 & 9 & \textit{Pool}     & 0.0330 & 0.4348 & 0.0242 \\
\midrule
3 & 10 & \textit{Local} & 0.0267 & 0.1492 & 0.0153 \\
3 & 10 & \textit{Pool}     & 0.0322 & 0.4089 & 0.0229 \\
\midrule
\multicolumn{3}{l}{VGGT \textit{(Base Model)}} & \textcolor{gray}{0.0374} & \textcolor{gray}{0.2934}& \textcolor{gray}{0.0186} \\
\bottomrule[0.17em]
\end{tabular}
}
\label{tab:vggt_pool_strategy}
\end{table}

\subsection{Layer partitioning with entropy thresholds}

To determine the thresholds for intra-frame selection strategies, we introduce two layer indices, $l_{\rm local}$ and $l_{\rm sample}$, which partition the global attention layers of visual geometry transformers into groups adopting different intra-frame strategies. A natural question to ask is that, \textit{can these partitioning rules can be made adaptive to the input sequence?}

Since the choice of $l_{\rm local}$ and $l_{\rm sample}$ is motivated by observations of the entropy of global attention patterns, a straightforward extension is to use entropy on each layer to adaptively group layers for different intra-frame strategy. Concretely, we define two entropy thresholds, $\tau_1$ and $\tau_2$. Starting from the first layer, we apply the most aggressive strategy, which is to replace global attention with local attention, until reaching the first layer where $\mathcal{H} < \tau_1$. From that point, we switch to the \textit{Standard} intra-frame downsampling strategy, and continue until encountering the first layer where the inter-frame entropy drops to $\mathcal{H} < \tau_2$.

The results are reported in~\Cref{tab:tau_ablation_1} and~\Cref{tab:tau_ablation_2}. While this adaptive scheme achieves competitive performance and serves as a viable alternative for determining layer-wise strategies, we do not adopt it in our final model for reasons similar to the \textit{TLD} strategy in~\Cref{sec:tld}. Specifically, computing attention entropy on-the-fly during the global attention pass introduces significant overhead, adding approximately 7 seconds for a 500-frame scene. Therefore, although promising, we present this approach as a reference to motivate future work on more efficient ways of leveraging entropy to inform intra-frame strategy selection.

\begin{table}[ht]
\centering
\rowcolors{2}{white}{uoftcoolgray!25}
\caption{Quantitative results of using entropy thresholds to determine the intra-frame strategies for each layer. Experiments are conducted on camera pose estimation with $K=25$ and $\sigma=3$ on Neural RGB-D~\cite{Azinovic_2022_CVPR}.}
\vspace{5pt}
\resizebox{0.65\linewidth}{!}{
\begin{tabular}{ccccc}
\toprule[0.17em]
\textbf{$\tau_1$} & \textbf{$\tau_2$} & \textbf{ATE ($\downarrow$)} & \textbf{RPE-rot ($\downarrow$)} & \textbf{RPE-trans ($\downarrow$)} \\
\midrule
0.99 & 0.95 & 0.0278 & 0.2948 & 0.0185 \\
0.99 & 0.92 & 0.0279 & 0.3455 & 0.0198 \\
0.97 & 0.95 & 0.0271 & 0.1871 & 0.0165 \\
0.97 & 0.92 & 0.0270 & 0.2405 & 0.0176 \\
0.97 & 0.90 & 0.0263 & 0.1523 & 0.0155 \\
0.95 & 0.92 & 0.0264 & 0.1911 & 0.0164 \\
0.95 & 0.90 & 0.0262 & 0.1469 & 0.0153 \\
0.95 & 0.88 & 0.0261 & 0.1455 & 0.0153 \\
0.93 & 0.90 & 0.0273 & 0.1497 & 0.0156 \\
0.93 & 0.88 & 0.0271 & 0.1491 & 0.0155 \\
0.91 & 0.88 & 0.0444 & 0.2820 & 0.0223 \\
\midrule
\multicolumn{2}{l}{\textbf{\methodname (Ours)}} & 0.0270 & 0.2409 & 0.0176 \\
\midrule
\multicolumn{2}{l}{VGGT \textit{(Base Model)}} & \textcolor{gray}{0.0374} & \textcolor{gray}{0.2934} & \textcolor{gray}{0.0186} \\
\bottomrule[0.17em]
\end{tabular}
}
\label{tab:tau_ablation_1}
\end{table}

\begin{table}[ht]
\centering
\rowcolors{2}{white}{uoftcoolgray!25}
\caption{Quantitative results of using entropy thresholds to determine the intra-frame strategies for each layer. Experiments are conducted on camera pose estimation with $K=25$ and $\sigma=2$ on 7-Scenes~\cite{7scenes}.}
\vspace{5pt}
\resizebox{0.65\linewidth}{!}{
\begin{tabular}{ccccc}
\toprule[0.17em]
\textbf{$\tau_1$} & \textbf{$\tau_2$} & \textbf{ATE ($\downarrow$)} & \textbf{RPE-rot ($\downarrow$)} & \textbf{RPE-trans ($\downarrow$)} \\
\midrule
0.99 & 0.95 & 0.0675 & 0.4446 & 0.0168 \\
0.99 & 0.92 & 0.0675 & 0.4428 & 0.0167 \\
0.97 & 0.95 & 0.0672 & 0.4471 & 0.0164 \\
0.97 & 0.92 & 0.0671 & 0.4454 & 0.0164 \\
0.97 & 0.90 & 0.0673 & 0.4471 & 0.0165 \\
0.95 & 0.92 & 0.0672 & 0.4454 & 0.0164 \\
0.95 & 0.90 & 0.0673 & 0.4471 & 0.0165 \\
0.95 & 0.88 & 0.0682 & 0.4461 & 0.0161 \\
0.93 & 0.90 & 0.0673 & 0.4517 & 0.0165 \\
0.93 & 0.88 & 0.0682 & 0.4506 & 0.0161 \\
0.91 & 0.88 & 0.0715 & 0.4880 & 0.0176 \\
\midrule
\multicolumn{2}{l}{\textbf{\methodname (Ours)}} & 0.0673 & 0.4471 & 0.0165 \\
\midrule
\multicolumn{2}{l}{VGGT \textit{(Base Model)}} & \textcolor{gray}{0.0698} & \textcolor{gray}{0.4953} & \textcolor{gray}{0.0178} \\
\bottomrule[0.17em]
\end{tabular}
}
\label{tab:tau_ablation_2}
\end{table}

\subsection{Intra-frame strategy for late layers}
Observing from the attention pattern analysis in~\Cref{fig:attention_pattern}, we notice that several of the latest layers also have the diluted attention patterns, which motivates us to consider: \textit{can intra-frame selection be applied on these final layers as well?} To investigate on this question, we introduce another threshold $l_{\rm late}$, such that intra-frame downsampling is also applied to layers with indices $l \geqslant l_{\rm late}$. Results are presented in~\Cref{tab:ab_layer_sigma_7scenes} and~\Cref{tab:ab_layer_sigma_nrgbd}. While performance can improve in certain cases, it is generally more sensitive to the choice of $l_{\rm late}$, not as robust as to $l_{\rm local}$ and $l_{\rm sample}$ as shown in~\Cref{tab:ab_intra_frame} and~\Cref{tab:ab_intra_frame_vggt}. We attribute this sensitivity to the proximity of these layers to the final output, where small changes can have a large impact on the final performance.



\begin{table}[ht]
\rowcolors{2}{white}{uoftcoolgray!25}
\centering
\captionof{table}{Quantitative results on applying intra-frame downsampling on selected late layers determined by $l_{\rm late}$. The experiments are conducted with $K=25$ on the camera pose estimation on 7-Scenes~\cite{7scenes}. $l_{\rm late}$ demonstrates inferior robustness than the choices of $l_{\rm local}$ and $l_{\rm sample}$.}
\vspace{5pt}
\resizebox{0.7\linewidth}{!}{
    \begin{tabular}{ccccc}
    \toprule[0.17em]
    $\sigma$ & $l_{\rm late}$ &
    \textbf{ATE ($\downarrow$)} &
    \textbf{RPE-rot ($\downarrow$)} &
    \textbf{RPE-trans ($\downarrow$)} \\
    \midrule
    \multicolumn{2}{l}{VGGT \textit{(Base Model)}} &\textcolor{gray}{0.0698} & \textcolor{gray}{0.4953}& \textcolor{gray}{0.0178}\\
    \midrule
    2 & 18 & 0.0718 & 0.4496 & 0.0166 \\
    2 & 19 & 0.0687 & 0.4448 & 0.0165 \\
    2 & 20 & 0.0686 & 0.4460 & 0.0165 \\
    2 & 21 & 0.0681 & 0.4453 & 0.0165 \\
    2 & 22 & 0.0683 & 0.4458 & 0.0165 \\
    2 & 23 & 0.0683 & 0.4462 & 0.0165 \\
    \midrule
    3 & 18 & 0.0758 & 0.4585 & 0.0171 \\
    3 & 19 & 0.0702 & 0.4481 & 0.0167 \\
    3 & 20 & 0.0710 & 0.4494 & 0.0167 \\
    3 & 21 & 0.0699 & 0.4471 & 0.0167 \\
    3 & 22 & 0.0699 & 0.4473 & 0.0167 \\
    3 & 23 & 0.0698 & 0.4478 & 0.0166 \\
    \bottomrule[0.17em]
    \end{tabular}
}
\label{tab:ab_layer_sigma_7scenes}
\end{table}

\begin{table}[ht]
\rowcolors{2}{white}{uoftcoolgray!25}
\centering
\captionof{table}{Quantitative results on applying intra-frame downsampling on selected late layers determined by $l_{\rm late}$. The experiments are conducted with $K=25$ on the camera pose estimation on Neural RGB-D~\cite{Azinovic_2022_CVPR}. $l_{\rm late}$ demonstrates inferior robustness than the choices of $l_{\rm local}$ and $l_{\rm sample}$.}
\vspace{5pt}
\resizebox{0.7\linewidth}{!}{
    \begin{tabular}{ccccc}
    \toprule[0.17em]
    $\sigma$ & $l_{\rm late}$ &
    \textbf{ATE ($\downarrow$)} &
    \textbf{RPE-rot ($\downarrow$)} &
    \textbf{RPE-trans ($\downarrow$)} \\
    \midrule
    \multicolumn{2}{l}{VGGT \textit{(Base Model)}} & \textcolor{gray}{0.0374}& \textcolor{gray}{0.2934}& \textcolor{gray}{0.0186}\\
    \midrule
    2 & 18 & 0.0315 & 0.2008 & 0.0170 \\
    2 & 19 & 0.0286 & 0.1842 & 0.0162 \\
    2 & 20 & 0.0277 & 0.1823 & 0.0163 \\
    2 & 21 & 0.0274 & 0.1818 & 0.0163 \\
    2 & 22 & 0.0273 & 0.1816 & 0.0163 \\
    2 & 23 & 0.0271 & 0.1820 & 0.0163 \\
    \midrule
    3 & 18 & 0.0412 & 0.3229 & 0.0236 \\
    3 & 19 & 0.0321 & 0.2540 & 0.0180 \\
    3 & 20 & 0.0311 & 0.2519 & 0.0181 \\
    3 & 21 & 0.0300 & 0.2463 & 0.0180 \\
    3 & 22 & 0.0297 & 0.2457 & 0.0179 \\
    3 & 23 & 0.0293 & 0.2451 & 0.0179 \\
    \bottomrule[0.17em]
    \end{tabular}
}
\label{tab:ab_layer_sigma_nrgbd}
\end{table}



\section{Additional experimental results for sensitivity analysis}
\label{sec:additional_sensitivity}

\subsection{Threshold hyperparameters}
In~\Cref{sec:ablation}, we demonstrate that the threshold hyperparameters $l_{\rm local}$ and $l_{\rm sample}$ are robust when with $\pi^3$ as the base model. We further verify this robustness with VGGT as the base model in~\Cref{tab:ab_intra_frame_vggt}, further enhancing the reliability and soundness of our method.

\subsection{Performance under varying sequence lengths}

As the goal of our method is to accelerate feed-forward 3D reconstruction, with the primary part of our work demonstrating how the inference time of our model can grow near-linearly with minimized performance compromise. To better complement our evaluation with varying length sequences, we evaluate chronological prefixes of 100-400 frames with the same sampling interval of 2 as the main experiments in~\Cref{sec:compare_with_others}. After sampling, since two scenes in 7-Scenes only contain 250 frames, we average results over all 18 scenes for the 100/200-frame settings and over the 16 eligible scenes for the 300/400-frame settings. As shown in~\Cref{tab:sequence_length_comparison}, when the sequence length becomes shorter, our method can still remain competitive with the baseline and other training-free acceleration methods, although it may falls short to methods like LiteVGGT~\cite{shu2025litevggtboostingvanillavggt} which requires additional training for the accelerated model.

\begin{table}
\rowcolors{2}{white}{uoftcoolgray!25}
\centering
\captionof{table}{Sensitivity analysis on the layer partition thresholds $l_{\rm local}$ and $l_{\rm sample}$ for intra-frame selection with $K=25$ on 7-Scenes~\cite{7scenes} using VGGT, with the \textit{\textcolor{uoftsecondaryblue}{blue}} row indicating our current parameter choice. Performance remains stable across a wide range of threshold choices, demonstrating that our method is robust to these hyperparameter choices, as long as they are consistent with the observed layer-wise attention patterns.}
\vspace{5pt}
\resizebox{0.7\linewidth}{!}{
    \begin{tabular}{ccccc}
    \toprule[0.17em]
    $l_{\rm local}$  & $l_{\rm sample}$  & 
    \textbf{ATE ($\downarrow$)} & 
    \textbf{RPE-rot ($\downarrow$)} & 
    \textbf{RPE-trans ($\downarrow$)} \\ 
    \midrule
    \multicolumn{2}{l}{VGGT \textit{(Base Model)}} & \textcolor{gray}{0.0698}  & \textcolor{gray}{0.4953} & \textcolor{gray}{0.0178} \\
    \midrule
    1 & 8 & 0.0670 & 0.4433 & 0.0165\\
    1 & 9 & 0.0676 & 0.4476 & 0.0167\\
    1 & 10 & 0.0675 & 0.4494 & 0.0166\\
    2 & 8 & 0.0671 & 0.4454 & 0.0164\\
    \textcolor{uoftsecondaryblue}{2} & \textcolor{uoftsecondaryblue}{9} & \textcolor{uoftsecondaryblue}{0.0677} & \textcolor{uoftsecondaryblue}{0.4495} & \textcolor{uoftsecondaryblue}{0.0166}\\
    2 & 10 & 0.0677 & 0.4518 & 0.0165\\
    3 & 8 & 0.0672 & 0.4501 & 0.0164\\
    3 & 9 & 0.0678 & 0.4532 & 0.0166\\
    3 & 10 & 0.0680 & 0.4539 & 0.0166\\
    4 & 8 & 0.0674 & 0.4555 & 0.0167\\
    4 & 9 & 0.0680 & 0.4589 & 0.0169\\
    4 & 10 & 0.0681 & 0.4581 & 0.0167\\
    \bottomrule[0.17em]
    \end{tabular}
}
\label{tab:ab_intra_frame_vggt}
\end{table}

\begin{table}[ht]
\centering
\rowcolors{2}{white}{uoftcoolgray!25}
\caption{Comparison on camera pose estimation and inference time on 7-Scenes with varying sequence lengths. Inputs are chronological prefixes of the scenes. The two shorter sequences are excluded at 300/400 frames. Best and second-best accelerated methods are \textbf{bold} and \underline{underlined}. Our \methodname uses $K=25$, $\sigma=3$.}
\vspace{5pt}
\resizebox{0.8\linewidth}{!}{%
\begin{tabular}{lcccc}
\toprule[0.17em]
\textbf{Method} & \makecell{\textbf{ATE ($\downarrow$)}} & \makecell{\textbf{RPE-rot ($\downarrow$)}} & \makecell{\textbf{RPE-trans ($\downarrow$)}} & \makecell{\textbf{Time ($\downarrow$)}} \\
\midrule
\rowcolor{white}
\multicolumn{5}{c}{\textbf{100 frames}} \\
\midrule
VGGT \textit{(Base Model)} & \textcolor{gray}{0.0293} & \textcolor{gray}{0.3179} & \textcolor{gray}{0.0095} & \textcolor{gray}{8.12s} \\
FastVGGT & 0.0301 & \underline{0.3399} & \underline{0.0107} & 8.15s \\
SparseVGGT (SR: 50\%) & 0.0329 & 0.3683 & 0.0121 & 5.88s \\
SparseVGGT (SR: 75\%) & 0.0345 & 0.4596 & 0.0156 & 5.03s \\
Co-Me & 0.0367 & 0.4594 & 0.0187 & \underline{3.70} \\
LiteVGGT & \textbf{0.0285} & 0.3648 & 0.0122 & \textbf{3.02s} \\
\textbf{\methodname (Ours)} & \underline{0.0289} & \textbf{0.3223} & \textbf{0.0099} & 6.96s \\
\midrule
\rowcolor{white}
\multicolumn{5}{c}{\textbf{200 frames}} \\
\midrule
VGGT \textit{(Base Model)} & \textcolor{gray}{0.0563} & \textcolor{gray}{0.3919} & \textcolor{gray}{0.0131} & \textcolor{gray}{25.57s} \\
FastVGGT & 0.0582 & \underline{0.4055} & \underline{0.0140} & 18.51s \\
SparseVGGT (SR: 50\%) & 0.0624 & 0.4691 & 0.0164 & 15.81s \\
SparseVGGT (SR: 75\%) & 0.0659 & 0.5977 & 0.0218 & 12.25s \\
Co-Me & 0.0742 & 0.6494 & 0.0281 & \underline{10.38s} \\
LiteVGGT & \underline{0.0528} & 0.4344 & 0.0160 & \textbf{7.23s} \\
\textbf{\methodname (Ours)} & \textbf{0.0525} & \textbf{0.3780} & \textbf{0.0130} & 14.10s \\
\midrule
\rowcolor{white}
\multicolumn{5}{c}{\textbf{300 frames}} \\
\midrule
VGGT \textit{(Base Model)} & \textcolor{gray}{0.0651} & \textcolor{gray}{0.4898} & \textcolor{gray}{0.0147} & \textcolor{gray}{51.83s} \\
FastVGGT & 0.0687 & \underline{0.4585} & \underline{0.0153} & 31.58s \\
SparseVGGT (SR: 50\%) & 0.0733 & 0.5543 & 0.0176 & 30.48s \\
SparseVGGT (SR: 75\%) & 0.0829 & 0.7012 & 0.0237 & 22.26s \\
Co-Me & 0.0883 & 0.7937 & 0.0294 & \underline{20.27s} \\
LiteVGGT & \textbf{0.0590} & 0.4977 & 0.0175 & \textbf{12.84s} \\
\textbf{\methodname (Ours)} & \underline{0.0624} & \textbf{0.4450} & \textbf{0.0146} & 21.29s \\
\midrule
\rowcolor{white}
\multicolumn{5}{c}{\textbf{400 frames}} \\
\midrule
VGGT \textit{(Base Model)} & \textcolor{gray}{0.0724} & \textcolor{gray}{0.5066} & \textcolor{gray}{0.0165} & \textcolor{gray}{87.42s} \\
FastVGGT & 0.0763 & \underline{0.4627} & \underline{0.0163} & 46.62s \\
SparseVGGT (SR: 50\%) & 0.0862 & 0.5959 & 0.0192 & 49.29s \\
SparseVGGT (SR: 75\%) & 0.0932 & 0.7539 & 0.0256 & 34.36s \\
Co-Me & 0.0956 & 0.9304 & 0.0337 & 33.26s \\
LiteVGGT & \underline{0.0685} & 0.7061 & 0.0202 & \textbf{19.23s} \\
\textbf{\methodname (Ours)} & \textbf{0.0678} & \textbf{0.4482} & \textbf{0.0154} & \underline{28.49s} \\
\bottomrule[0.17em]
\end{tabular}%
}
\label{tab:sequence_length_comparison}
\end{table}

\section{Application on other architecture of visual geometry transformers}
\label{sec:other_vggt}

The visual geometry transformers, VGGT~\cite{wang2025vggtvisualgeometrygrounded} and $\pi^3$~\cite{wang2025pi3permutationequivariantvisualgeometry} that we mainly experiment with in the main paper, adopt global attention layers for cross-view information processing, which are also the major architecture that existing methods like FastVGGT~\cite{shen2025fastvggttrainingfreeaccelerationvisual}, SparseVGGT~\cite{wang2025sparsevggt} target to accelerate. To demonstrate the broad application of our method as a general plug-in solution for various types of visual geometry transformers, we additionally illustrate below that base models like Depth Anything 3~\cite{lin2026depth} which uses a different architecture (input-adaptive cross-view self-attention by token rearrangement) for cross-view information processing can also easily benefit from our solution. In~\Cref{tab:da3_large}, we show that our method can also effectively boost the inference speed for Depth Anything 3 without performance compromise. In contrast, existing methods like SparseVGGT that require inspection on global attention maps for token compression and merging are not very straightforward to be adapted to models like Depth Anything 3.

\begin{table}[ht]
\centering
\rowcolors{2}{white}{uoftcoolgray!25}
\caption{Application to Depth Anything 3 Large (DA3-Large) on 7-Scenes. \methodname uses $K=25$ and $\sigma=2$. Inference time is measured with 500-frame sequences.}
\vspace{5pt}
\resizebox{0.83\linewidth}{!}{%
\begin{tabular}{lcccc}
\toprule[0.17em]
\textbf{Method} & \textbf{ATE ($\downarrow$)} & \textbf{RPE-rot ($\downarrow$)} & \textbf{RPE-trans ($\downarrow$)} & \textbf{Time ($\downarrow$)} \\
\midrule
DA3-Large & 0.0692 & 0.4259 & \textbf{0.0136} & 96.5s \\
\textbf{\methodname on DA3-Large (Ours)} & \textbf{0.0677} & \textbf{0.4177} & 0.0139 & \textbf{29.2s} \\
\bottomrule[0.17em]
\end{tabular}}
\label{tab:da3_large}
\end{table}


\section{Detailed explanation on evaluation metrics}
\label{sec:evaluation_metrics}

We provide detailed explanations of the evaluation metrics in the three tasks: camera pose estimation, 3D reconstruction, and video depth estimation as below.

\subsection{Camera pose estimation}

\textbf{Absolute Trajectory Error (ATE).}
ATE measures the global error between the estimated trajectory and the ground truth trajectory. Given estimated poses $\{\hat{T}_i\}$ and ground truth poses $\{T_i\}$, we compute the Root Mean Square Error (RMSE) between aligned camera positions:
\begin{equation}
\text{ATE} = \sqrt{\frac{1}{N} \sum_{i=1}^{N} \| \mathbf{p}_i - \hat{\mathbf{p}}_i \|^2},
\end{equation}
where $\mathbf{p}_i$ and $\hat{\mathbf{p}}_i$ denote the ground truth and estimated camera centers after alignment. ATE captures long-term drift over the trajectory.

\textbf{Relative Pose Error (RPE).}
RPE evaluates local motion consistency over a fixed interval $\Delta$. In evaluation, we use interval $\Delta=1$, indicating that we report the relative pose error between adjacent frames. Let $T_{i,i+\Delta} = T_i^{-1} T_{i+\Delta}$ and $\hat{T}_{i,i+\Delta} = \hat{T}_i^{-1} \hat{T}_{i+\Delta}$. Each relative transformation $T_{i,i+\Delta} \in {\rm SE}(3)$ can be decomposed as:
\begin{equation}
T_{i,i+\Delta} =
\begin{bmatrix}
R_{i,i+\Delta} & \mathbf{t}_{i,i+\Delta} \\
0 & 1
\end{bmatrix},
\end{equation}
where $R_{i,i+\Delta} \in {\rm SO}(3)$ denotes the relative rotation matrix and $\mathbf{t}_{i,i+\Delta} \in \mathbb{R}^3$ denotes the relative translation vector. \textbf{RPE-rot} captures the rotation error of the relative pose:
\begin{equation}
\text{RPE-rot} = \frac{1}{N} \sum_{i} \angle \left( R_{i,i+\Delta}^{-1} \hat{R}_{i,i+\Delta} \right),
\end{equation}
where $\angle(\cdot)$ computes the rotation angle. \textbf{RPE-trans} captures the translation error of the relative pose:
\begin{equation}
\text{RPE-trans} = \frac{1}{N} \sum_{i} \| \mathbf{t}_{i,i+\Delta} - \hat{\mathbf{t}}_{i,i+\Delta} \|.
\end{equation}

\subsection{3D reconstruction}
\textbf{Accuracy (Acc).}
Accuracy measures how close the predicted point cloud $\hat{P}$ is to the ground truth point cloud $P$:
\begin{equation}
\text{Acc} = \frac{1}{|\hat{P}|} \sum_{\hat{p} \in \hat{P}} \min_{p \in P} \| \hat{p} - p \|.
\end{equation}
\textbf{Completeness (Comp).}
On the contrary to accuracy, completeness evaluates how well the ground truth geometry is covered by the prediction:
\begin{equation}
\text{Comp} = \frac{1}{|P|} \sum_{p \in P} \min_{\hat{p} \in \hat{P}} \| p - \hat{p} \|.
\end{equation}
\textbf{Normal Consistency (NC).}
Normal consistency measures the alignment between predicted and ground-truth surface normals. For each predicted point $\hat{p} \in \hat{P}$, we find its nearest neighbor $p \in P$ in the ground truth point cloud, and compute the cosine similarity between their normals:
\begin{equation}
\text{NC} = \frac{1}{|\hat{P}|} \sum_{\hat{p} \in \hat{P}} 
\left\langle \mathbf{n}(p), \mathbf{n}(\hat{p}) \right\rangle,
\end{equation}
where $\mathbf{n}(\hat{p})$ and $\mathbf{n}(p)$ denote the predicted and ground truth normals at $\hat{p}$ and its nearest neighbor $p$, respectively.

\subsection{Video depth estimation}

Let $d_i$ and $\hat{d}_i$ denote the ground-truth and predicted depths at pixel $i$, respectively, and $N$ be the number of valid pixels.

\textbf{Absolute Relative Error (Abs Rel).} Abs Rel measures the absolute error normalized by the ground-truth depth. It emphasizes relative accuracy, assigning larger penalties to errors at closer depths where precision is typically more important:
\begin{equation}
\text{Abs Rel} = \frac{1}{N} \sum_{i} \frac{|d_i - \hat{d}_i|}{d_i}.
\end{equation}
\textbf{Squared Relative Error (Sq Rel).} Sq Rel penalizes large errors more heavily due to the squared term. This makes it more sensitive to outliers and particularly useful for evaluating robustness when large depth errors occur:
\begin{equation}
\text{Sq Rel} = \frac{1}{N} \sum_{i} \frac{(d_i - \hat{d}_i)^2}{d_i}.
\end{equation}
\textbf{Root Mean Square Error (RMSE).} RMSE measures the absolute discrepancy in depth values and strongly penalizes large deviations. It reflects overall reconstruction fidelity in the original depth scale, but can be dominated by errors in distant regions where depth values are larger:
\begin{equation}
\text{RMSE} = \sqrt{\frac{1}{N} \sum_{i} (d_i - \hat{d}_i)^2}.
\end{equation}
\textbf{Log RMSE.} By operating in log space, Log RMSE reduces sensitivity to absolute scale and instead emphasizes relative differences. It is particularly suitable when the depth range is large, as it balances errors across near and far regions and better reflects perceptual quality:
\begin{equation}
\text{Log RMSE} = \sqrt{\frac{1}{N} \sum_{i} (\log d_i - \log \hat{d}_i)^2}.
\end{equation}
\textbf{Threshold Accuracy ($\delta < 1.25$).} This accuracy metric is the percentage of pixels satisfying $\delta_i < 1.25$, where
\begin{equation}
\delta_i = \max\left( \frac{d_i}{\hat{d}_i}, \frac{\hat{d}_i}{d_i} \right).
\end{equation}
This metric evaluates the fraction of predictions that fall within a multiplicative error bound of the ground truth. It provides an intuitive measure of reliability, indicating how many predictions are sufficiently accurate rather than averaging errors.

\section{Licenses for existing assets}
\label{sec:licenses}

The following list contains licenses for data and model used in the paper:

\begin{itemize}
    \item VGGT~\cite{wang2025vggtvisualgeometrygrounded}: View the link for details
    
    \url{https://github.com/facebookresearch/vggt}
    \item $\pi^3$~\cite{wang2025pi3permutationequivariantvisualgeometry}: BSD-3-Clause License
    
    \url{https://github.com/yyfz/Pi3}
    \item 7-Scenes~\cite{7scenes}: View the link for details
    
    \url{https://www.microsoft.com/en-us/research/project/rgb-d-dataset-7-scenes/}
    \item Neural RGB-D~\cite{Azinovic_2022_CVPR}: View the link for details
    
    \url{https://github.com/dazinovic/neural-rgbd-surface-reconstruction}
    \item Bonn~\cite{bonn}: Creative Commons Attribution-NonCommercial-ShareAlike 3.0 Unported License
    
    \url{https://www.ipb.uni-bonn.de/data/rgbd-dynamic-dataset/index.html}
    \item TUM-Dynamics~\cite{tum_dynamics}: CC-BY-4.0 License
    
    \url{https://cvg.cit.tum.de/data/datasets/rgbd-dataset}
\end{itemize}

The license names can also be found in the above links. As some datasets contain different licenses for different scenes, we do not list them here one by one to save space.

\section{Additional discussions}
\label{sec:additional_discussion}


Recently, concurrent works~\cite{zhang2026logerlongcontextgeometricreconstruction, elflein2026vggt3offlinefeedforward3d, jin2026zipmap, wang2026tttlrm, xie2026scal3rscalabletesttimetraining} have explored replacing the quadratic-time attention with test-time training (TTT)~\cite{sun2025ttt} layers to achieve linear-time scaling in visual geometry transformers. While these approaches substantially improve inference efficiency, they typically incur a performance compromise compared to standard attention-based architectures and require considerable computational resources to train the TTT layers. In contrast, our method is fully \textit{training-free} and can be readily applied as a plug-in to a wide range of existing models. Therefore, our approach is complementary to these methods and can potentially be integrated with them for further efficiency improvements, and we leave this as an interesting future work.




\section{Societal impact}
\label{sec:Societal}

We expect our work to have an overall positive societal impact. By accelerating visual geometry transformers, our method improves the accessibility of advanced 3D reconstruction technologies, enabling individual researchers and smaller organizations with limited computational resources to utilize these visual geometry transformers more effectively. In this sense, improving efficiency can contribute to a more equitable and democratized AI ecosystem~\cite{leblanc2026distill3rpipelinedemocratizing3d}, where high-quality 3D reconstruction systems are not exclusively accessible to institutions with extensive computational infrastructure.

More efficient visual geometry transformers may also benefit a broad range of downstream applications with positive societal value, such as digital cultural heritage preservation. In particular, reducing inference cost facilitates deployment on resource-constrained platforms and edge devices, enabling efficient deployment that were previously impractical due to computational limitations.

\textbf{Potential negative societal impact.} We do not identify obvious negative societal impact from our work. Potential risks may include misuse for malicious purposes such as unlawful surveillance that harms individual privacy.